\batchmode
\makeatletter
\def\input@path{{\string"E:/Trabajo Angel/Mis articulos/Finished/Trajectory PHD-CPHD filter/Accepted/\string"}}
\makeatother
\documentclass[english]{IEEEtran}
\usepackage[T1]{fontenc}
\usepackage[latin9]{inputenc}
\usepackage{float}
\usepackage{amsmath}
\usepackage{amsthm}
\usepackage{amssymb}
\usepackage{stmaryrd}
\usepackage{graphicx}
\PassOptionsToPackage{normalem}{ulem}
\usepackage{ulem}

\makeatletter

\providecommand{\tabularnewline}{\\}
\floatstyle{ruled}
\newfloat{algorithm}{tbp}{loa}
\providecommand{\algorithmname}{Algorithm}
\floatname{algorithm}{\protect\algorithmname}

\theoremstyle{plain}
\newtheorem{thm}{\protect\theoremname}
\theoremstyle{definition}
\newtheorem{example}[thm]{\protect\examplename}
\theoremstyle{plain}
\newtheorem{prop}[thm]{\protect\propositionname}

\usepackage{cite} 
\usepackage[margin=8pt,font=footnotesize]{caption}
\usepackage{algorithm}
\usepackage{algpseudocode}

\usepackage{amsmath}  
\allowdisplaybreaks

\makeatother

\usepackage{babel}
\providecommand{\examplename}{Example}
\providecommand{\propositionname}{Proposition}
\providecommand{\theoremname}{Theorem}

\begin{document}

\title{Trajectory PHD and CPHD filters}

\author{Ángel F. García-Fernández, Lennart Svensson\thanks{A. F. García-Fernández is with the Department of Electrical Engineering and Electronics, University of Liverpool, Liverpool L69 3GJ, United Kingdom (email: angel.garcia-fernandez@liverpool.ac.uk). L. Svensson is with the Department of Electrical Engineering, Chalmers University of Technology, SE-412 96 Gothenburg, Sweden (email: lennart.svensson@chalmers.se).}}
\maketitle
\begin{abstract}
This paper presents the probability hypothesis density filter (PHD)
and the cardinality PHD (CPHD) filter for sets of trajectories, which
are referred to as the trajectory PHD (TPHD) and trajectory CPHD (TCPHD)
filters. Contrary to the PHD/CPHD filters, the TPHD/TCPHD filters
are able to produce trajectory estimates from first principles. The
TPHD filter is derived by recursively obtaining the best Poisson multitrajectory
density approximation to the posterior density over the alive trajectories
by minimising the Kullback-Leibler divergence. The TCPHD is derived
in the same way but propagating an independent identically distributed
(IID) cluster multitrajectory density approximation. We also propose
the Gaussian mixture implementations of the TPHD and TCPHD recursions,
the Gaussian mixture TPHD (GMTPHD) and the Gaussian mixture TCPHD
(GMTCPHD), and the $L$-scan computationally efficient implementations,
which only update the density of the trajectory states of the last
$L$ time steps.
\end{abstract}


\section{Introduction}

The probability hypothesis density (PHD) and cardinality PHD (CPHD)
filters are widely used random finite set (RFS) algorithms for multitarget
filtering, which aims to estimate the state of the targets at the
current time based on a sequence of measurements \cite{Mahler_book14,Granstrom12,Vo06,Lundquist13,Vo07,Whiteley10,Erdinc09}.
These filters have been successfully used in different applications
such as multitarget tracking \cite{Mahler_book14}, distributed multi-sensor
fusion \cite{Uney13,Battistelli13}, robotics \cite{Lee14,Mullane11},
computer vision \cite{Maggio08,Wood12}, road mapping \cite{Lundquist11}
and sensor control \cite{Ristic11b}.

The PHD/CPHD filters fit into the assumed density filtering framework
and propagate a  certain type of multitarget density on the current
set of targets through the prediction and update steps \cite{Angel15_d}.
The PHD filter considers a Poisson multitarget density, in which the
cardinality of the set is Poisson distributed and, for each cardinality,
its elements are independent and identically distributed (IID). On
the other hand, the CPHD filter considers an IID cluster multitarget
density, in which the cardinality distribution of the set is arbitrary
and, for each cardinality, its elements are IID. If the output of
either the prediction or the update step is no longer Poisson/IID
cluster, the PHD/CPHD filters obtain the best Poisson/IID cluster
approximation by minimising the Kullback-Leibler divergence (KLD).

The most important benefit of the PHD/CPHD filters is their low computational
burden, as they avoid the measurement-to-target association problem.
However, their main drawbacks are their relatively low performance
in some scenarios \cite{Mahler_book14,Franken09} and the fact that
they do not build tracks, which denote sequences of target states
that belong to the same target. The smoother versions of these filters
\cite{Mahler_book14,Nadarajah11,Nagappa17} do not solve these drawbacks. 

Despite the fact that the PHD/CPHD filters are unable to provide tracks
in a mathematically rigorous way, several track building procedures
have been proposed \cite{Lin06,Panta07,Panta09,Pollard11,Lu18}. A
track building procedure for PHD/CPHD filters was proposed in \cite{Lu17}
by adding labels \cite{Angel13,Vo13} to the target states. Nonetheless,
in the resulting labelled Poisson and labelled IID cluster densities,
there is total confusion in the label-to-target association so they
are not useful for track formation \cite[Sec. III.B]{Lu17}\cite[Sec. II.B]{Angel15_prov}.
To solve this issue in \cite{Lu17}, apart from the unique labels,
unique tags are added to the PHD components, as in \cite{Panta09},
and the original PHD/CPHD recursions are applied. However, in the
considered posterior density, the tags are not part of the target
state and are marginalised out. Therefore, the posterior is still
distributed as labelled Poisson or labelled IID cluster and, theoretically,
it does not have information to infer tracks. While tagging PHD components
works well in some scenarios, each PHD component does not generally
represent information about a unique target, as the corresponding
number of targets is Poisson distributed. In fact, adding tags to
the PHD components and reporting estimates with unique tags to build
trajectories, can lead to track switches, missed detections and false
targets when there is more than one target represented by the same
tag.

In this paper, we address the intrinsic inability of standard PHD/CPHD
filters to infer trajectories by developing PHD/CPHD filters that
provide tracks from first principles, without adding labels or tags.
We propose the trajectory PHD (TPHD) and trajectory CPHD (TCPHD) filters,
which follow the same assumed density filtering scheme as the PHD/CPHD
filters \cite{Angel18_c} with a fundamental difference: instead of
using a set of targets as the state variable, they use a set of trajectories
\cite{Svensson14,Angel15_prov}.  

\begin{figure}
\begin{centering}
\includegraphics[scale=0.6]{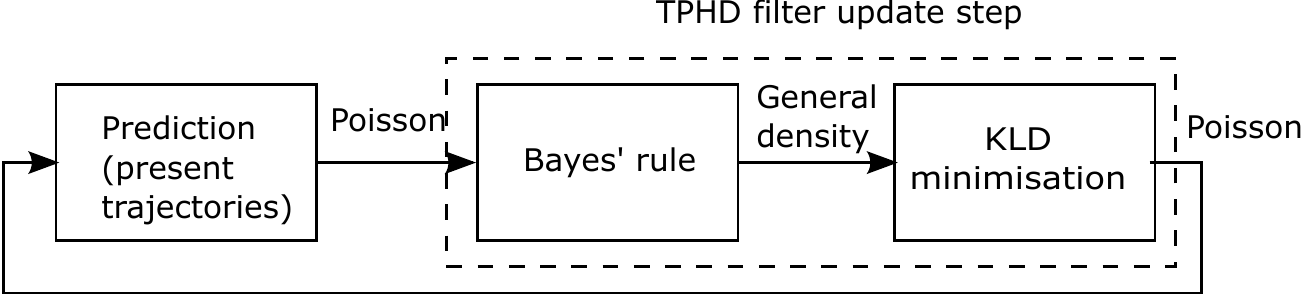}
\par\end{centering}
\caption{\label{fig:TPHD-filter-diagram.}TPHD filter diagram for estimating
the present trajectories at the current time. The TPHD filter assumes
that the multitrajectory densities involved are Poisson (on the space
of sets of trajectories). The output of Bayes' rule is no longer Poisson
but, in order to be able to perform the Bayesian recursion, it obtains
the best Poisson approximation to the filtering density by minimising
the KLD.}
\end{figure}

\begin{figure*}
\begin{centering}
\includegraphics[scale=0.6]{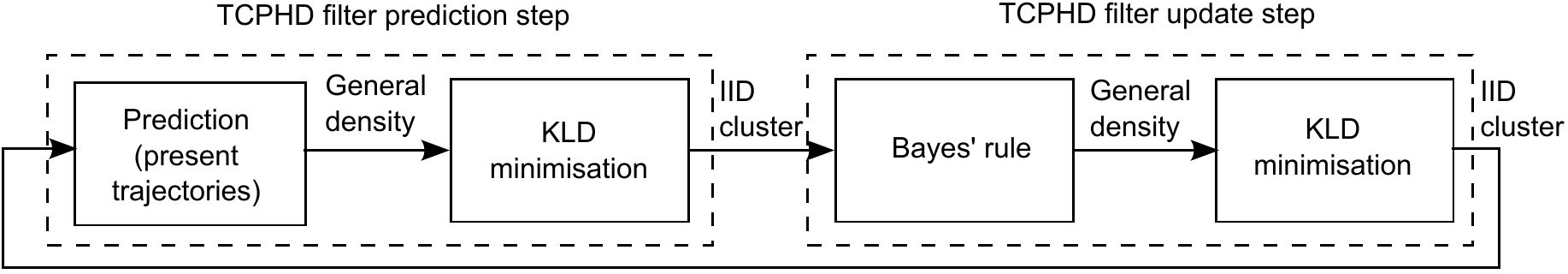}
\par\end{centering}
\caption{\label{fig:TCPHD-filter-diagram.}TCPHD filter diagram for estimating
the present trajectories at the current time. The TCPHD filter assumes
that the multitrajectory densities involved are IID cluster. The output
of the prediction and Bayes' rule are no longer IID clusters but the
TCPHD filter obtains the best IID cluster approximation by minimising
the KLD.}
\end{figure*}

The TPHD filter propagates a Poisson multitrajectory density on the
space of sets of trajectories through the prediction and update steps,
with a KLD minimisation after the update step. A diagram of the resulting
Bayesian recursion is given in Figure \ref{fig:TPHD-filter-diagram.}.
Similarly, the TCPHD filter propagates an IID cluster multitrajectory
density and performs a KLD minimisation after the prediction and update
steps, see Figure \ref{fig:TCPHD-filter-diagram.}. Due to the widespread
use of PHD/CPHD filters, this paper covers an important gap in the
literature, as we show how PHD/CPHD filtering can be endowed with
the ability to infer trajectories in a rigorous and principled way.
In particular, the TPHD and TCPHD filters are able to estimate the
trajectories of the alive targets by propagating a Poisson and an
IID cluster multitrajectory density through the filtering recursion
using KLD minimisations. The TPHD and TCPHD filters also avoid the
above-mentioned drawbacks of trajectory estimation based on labelling/tagging
the PHD. Apart from theoretically sound track formation, the proposed
filters also have the advantage, compared to previous track building
procedures used in PHD/CPHD filters, that they can update the information
regarding past states of the trajectories. 

In this paper, we also propose Gaussian mixture implementations of
the TPHD/TCPHD filters, which follow the spirit of the Gaussian mixture
PHD/CPHD filters \cite{Vo06,Vo07}. The resulting Gaussian mixture
TPHD (GMTPHD) and TCPHD (GMTCPHD) filters build trajectories under
a Poisson or IID cluster approximation, whose PHD is represented by
a Gaussian mixture. In this setting, a Gaussian component of the GMTPHD/GMTCPHD
filter represents information over entire trajectories, while a Gaussian
component in the Gaussian mixture PHD/CPHD filters (tagged or not)
represents information over current target states. It is therefore
straightforward to extract trajectory estimates from the GMTPHD/GMTCPHD
filters. Additionally, we propose a version of the GMTPHD/GMTCPHD
filters with lower computational burden called the $L$-scan GMTPHD/GMTCPHD
filters. In practice, these filters only update the multitrajectory
density of the trajectory states of the last $L$ time instant leaving
the rest unaltered, which is quite efficient for implementation. The
theoretical foundation of the $L$-scan GMTPHD filter is also based
on the assumed density filtering framework and KLD minimisations.
Preliminary results of this paper covering the TPHD filter were presented
in \cite{Angel18_c}. 

The remainder of the paper is organised as follows. Section \ref{sec:Background}
presents background material on sets of trajectories. In Section \ref{sec:Poisson-and-IID-RFS},
we introduce the Poisson and IID cluster multitrajectory densities
and some of their properties. The TPHD and TCPHD filters are derived
in Sections \ref{sec:Trajectory-PHD-filter} and \ref{sec:Trajectory-CPHD-filter},
respectively. Their Gaussian mixture implementations are provided
in Section \ref{sec:Gaussian-mixture-implementations}. Simulation
results are provided in Section \ref{sec:Simulations}. Finally, conclusions
are drawn in Section \ref{sec:Conclusions}.

\section{Background\label{sec:Background}}

In this section, we describe some background material on multiple
target tracking using sets of trajectories \cite{Angel15_prov}. We
review the considered variables, the set integral and cardinality
distribution for sets of trajectories in Sections \ref{subsec:Variables},
\ref{subsec:Set-integral} and \ref{subsec:Cardinality-distribution},
respectively. Finally, we introduce the PHD for sets of trajectories
in Section \ref{subsec:Probability-hypothesis-density}.

\subsection{Variables\label{subsec:Variables}}

A single target state $x\in\mathbb{R}^{n_{x}}$ contains information
of interest about the target, e.g., its position and velocity. A set
of single target states $\mathbf{x}$ belongs to $\mathcal{F}\left(\mathbb{R}^{n_{x}}\right)$
where $\mathcal{F}\left(\mathbb{R}^{n_{x}}\right)$ denotes the set
of all finite subsets of $\mathbb{R}^{n_{x}}$. We are interested
in estimating all target trajectories, where a trajectory consists
of a sequence of target states that can start at any time step and
end any time later on. Mathematically, a trajectory is represented
as a variable $X=\left(t,x^{1:i}\right)$ where $t$ is the initial
time step of the trajectory, $i$ is its length and $x^{1:i}=\left(x^{1},...,x^{i}\right)$
denotes a sequence of length $i$ that contains the target states
at consecutive time steps of the trajectory. 

We consider trajectories up to the current time step $k$. As a trajectory
$\left(t,x^{1:i}\right)$ exists from time step $t$ to $t+i-1$,
variable $\left(t,i\right)$ belongs to the set $I_{(k)}=\left\{ \left(t,i\right):0\leq t\leq k\,\mathrm{and}\,1\leq i\leq k-t+1\right\} $.
A single trajectory $X$ up to time step $k$ therefore belongs to
the space $T_{\left(k\right)}=\uplus_{\left(t,i\right)\in I_{(k)}}\left\{ t\right\} \times\mathbb{R}^{in_{x}}$,
where $\uplus$ stands for disjoint union, which is used to highlight
that the sets are disjoint. Similarly to the set ${\bf x}$ of targets,
we denote a set of trajectories up to time step $k$ as $\mathbf{X}\in\mathcal{F}\left(T_{\left(k\right)}\right)$. 

Given a trajectory $X=\left(t,x^{1:i}\right)$, the set $\tau^{k'}\left(X\right)$,
which can be empty, denotes the corresponding target state at a time
step $k'$. Given a set $\mathbf{X}$ of trajectories, the set $\tau^{k'}\left(\mathbf{X}\right)$
of target states at time $k'$ is $\tau^{k'}\left(\mathbf{X}\right)=\bigcup_{X\in\mathbf{X}}\tau^{k'}\left(X\right)$.

\subsection{Set integral\label{subsec:Set-integral}}

Given a real-valued function $\pi\left(\cdot\right)$ on the single
trajectory space $T_{\left(k\right)}$, its integral is \cite{Angel15_prov}
\begin{align}
\int\pi\left(X\right)dX & =\sum_{\left(t,i\right)\in I_{(k)}}\int\pi\left(t,x^{1:i}\right)dx^{1:i}.\label{eq:single_trajectory_integral}
\end{align}
This integral goes through all possible start times, lengths and target
states of the trajectory. Given a real-valued function $\pi\left(\cdot\right)$
on the space $\mathcal{F}\left(T_{\left(k\right)}\right)$ of sets
of trajectories, its set integral is \cite{Angel15_prov}
\begin{align}
\int\pi\left(\mathbf{X}\right)\delta\mathbf{X} & =\sum_{n=0}^{\infty}\frac{1}{n!}\int\pi\left(\left\{ X_{1},...,X_{n}\right\} \right)dX_{1:n}\label{eq:set_integral_trajectory}
\end{align}
where $X_{1:n}=\left(X_{1},...,X_{n}\right)$. A function $\pi\left(\cdot\right)$
is a multitrajectory density if $\pi\left(\cdot\right)\geq0$ and
its set integral is one.

\subsection{Cardinality distribution\label{subsec:Cardinality-distribution}}

Given a multitrajectory density $\pi\left(\cdot\right)$, its cardinality
distribution is 
\begin{align}
\rho_{\pi}\left(n\right) & =\frac{1}{n!}\int\pi\left(\left\{ X_{1},...,X_{n}\right\} \right)dX_{1:n},\label{eq:cardinality_distribution_trajectories}
\end{align}
which is analogous to the case where there is a set of targets.

\subsection{Probability hypothesis density\label{subsec:Probability-hypothesis-density}}

The PHD \cite{Mahler_book14} of a multitrajectory density $\pi\left(\cdot\right)$
is 
\begin{align}
D_{\pi}(X) & =\int\pi\left(\left\{ X\right\} \cup\mathbf{X}\right)\delta\mathbf{X}.\label{eq:PHD}
\end{align}
As in the PHD for RFS of targets, integrating the PHD in a region
$A\subseteq T_{\left(k\right)}$ gives us the expected number of trajectories
in this region \cite[Eq. (4.76)]{Mahler_book14}:
\begin{align}
\hat{N}_{A} & =\int_{A}D_{\pi}(X)dX\nonumber \\
 & =\sum_{\left(t,i\right)\in I_{(k)}}\int1_{A}\left(t,x^{1:i}\right)D_{\pi}(t,x^{1:i})dx^{1:i}\label{eq:expected_trajectory_number}
\end{align}
where $1_{A}\left(\cdot\right)$ is the indicator function of a subset
$A$: $1_{A}\left(z\right)=1$ if $z\in A$ and $1_{A}\left(z\right)=0$
otherwise. Therefore, the expected number of trajectories up to time
step $k$ is given by substituting $A=T_{\left(k\right)}$ into (\ref{eq:expected_trajectory_number}).
\begin{example}
\label{exa:PHD}Let us consider a multitrajectory density $\nu\left(\cdot\right)$
with PHD
\begin{align}
D_{\nu}\left(1,x^{1}\right) & =\mathcal{N}\left(x^{1};10,1\right)+\mathcal{N}\left(x^{1};1000,1\right)\label{eq:PHD_example1}\\
D_{\nu}\left(1,x^{1:2}\right) & =\mathcal{N}\left(x^{1:2};\left(10,10.1\right),\left[\begin{array}{cc}
1 & 1\\
1 & 2
\end{array}\right]\right),\label{eq:PHD_example2}
\end{align}
and $D_{\nu}\left(X\right)=0$ for $X\neq\left(1,x^{1}\right)$ and
$X\neq\left(1,x^{1:2}\right)$, where $\mathcal{N}\left(\cdot;m,P\right)$
is a Gaussian density with mean $m$ and covariance matrix $P$. The
expected number of trajectories that start at time one with length
1 is given by substituting $A=\left\{ 1\right\} \times\mathbb{R}^{n_{x}}$
into (\ref{eq:expected_trajectory_number}) so
\begin{align*}
\hat{N}_{A} & =\int D_{\nu}\left(1,x^{1}\right)dx^{1}=2.
\end{align*}
The expected number of trajectories up to time step $k=2$ is $\hat{N}_{T_{\left(k\right)}}=3$.
$\oblong$
\end{example}

\section{Poisson and IID cluster trajectory RFSs\label{sec:Poisson-and-IID-RFS}}

In this section, we explain the Poisson and IID cluster trajectory
RFSs.

\subsection{Multitrajectory densities}

\subsubsection{Poisson RFS}

For a Poisson RFS, the cardinality of the set is Poisson distributed
and, for each cardinality, its elements are IID. A Poisson multitrajectory
density $\nu\left(\cdot\right)$ has the form
\begin{align}
\nu\left(\left\{ X_{1},...,X_{n}\right\} \right) & =e^{-\lambda_{\nu}}\lambda_{\nu}^{n}\prod_{j=1}^{n}\breve{\nu}\left(X_{j}\right)\label{eq:Poisson_prior}
\end{align}
where $\breve{\nu}\left(\cdot\right)$ is a single trajectory density,
which implies 
\begin{align}
\int\breve{\nu}\left(X\right)dX & =1,\label{eq:integral_single_trajectory_pdf}
\end{align}
and $\lambda_{\nu}\geq0$. A Poisson multitrajectory density is characterised
by either its PHD $D_{\nu}(X)=\lambda_{\nu}\breve{\nu}\left(X\right)$
or by $\lambda_{\nu}$ and $\breve{\nu}\left(\cdot\right)$ \cite{Mahler_book14}.
As a result, using (\ref{eq:expected_trajectory_number}), the expected
number of trajectories is $\hat{N}_{T_{\left(k\right)}}=\lambda_{\nu}$.
Further, its cardinality distribution is given by 
\begin{align}
\rho_{\nu}\left(n\right)=\frac{1}{n!}\int\nu\left(\left\{ X_{1},...,X_{n}\right\} \right)dX_{1:n} & =\frac{1}{n!}e^{-\lambda_{\nu}}\lambda_{\nu}^{n}.\label{eq:Poisson_cardinality}
\end{align}
\begin{example}
We consider a Poisson RFS with the PHD of Example \ref{exa:PHD}.
Using (\ref{eq:Poisson_cardinality}), its cardinality distribution
is Poisson with $\lambda_{\nu}=3$ and, therefore, its single trajectory
density is $\breve{\nu}\left(X\right)=D_{\nu}\left(X\right)/3$. $\oblong$
\end{example}

\subsubsection{IID cluster RFS}

For an IID cluster RFS with multitrajectory density $\nu\left(\cdot\right)$,
the cardinality is distributed according to the probability mass function
$\rho_{\nu}\left(\cdot\right)$ and, for each cardinality, its elements
are IID according to a single trajectory density $\breve{\nu}\left(\cdot\right)$.
The resulting multitrajectory density is 
\begin{align}
\nu\left(\left\{ X_{1},...,X_{n}\right\} \right) & =\rho_{\nu}\left(n\right)n!\prod_{j=1}^{n}\breve{\nu}\left(X_{j}\right).\label{eq:iidc_prior}
\end{align}
As $\breve{\nu}\left(\cdot\right)$ is a single trajectory density,
it meets (\ref{eq:integral_single_trajectory_pdf}). The PHD of (\ref{eq:iidc_prior})
is given by \cite{Mahler_book14}
\begin{align}
D_{\nu}\left(x\right) & =\breve{\nu}\left(x\right)\sum_{n=0}^{\infty}n\rho_{\nu}\left(n\right)\label{eq:PHD_IID_cluster}
\end{align}
where the second factor corresponds to the expected number of trajectories.
An IID cluster density can be characterised either by $\rho_{\nu}\left(\cdot\right)$
and $\breve{\nu}\left(\cdot\right)$, or by $\rho_{\nu}\left(\cdot\right)$
and $D_{\nu}\left(\cdot\right)$. How to draw samples from an IID
cluster trajectory RFS, which includes the Poisson trajectory RFS
as a particular case, is explained in Appendix \ref{sec:Appendix_sampling}
in the supplementary material.

\subsection{KLD minimisation}

In this subsection, we provide two KLD minimisation theorems for Poisson
and IID cluster multitrajectory densities, which will be used to derive
the trajectory PHD/CPHD filters. 

The KLD $\mathrm{D}\left(\pi\left\Vert \nu\right.\right)$ between
multitrajectory densities $\pi\left(\cdot\right)$ and $\nu\left(\cdot\right)$
is given by \cite{Mahler_book14}
\begin{align}
\mathrm{D}\left(\pi\left\Vert \nu\right.\right) & =\int\pi\left(\mathbf{X}\right)\log\frac{\pi\left(\mathbf{X}\right)}{\nu\left(\mathbf{X}\right)}\delta\mathbf{X}.\label{eq:KLD_definition}
\end{align}
Then, the following theorems hold:
\begin{thm}
\label{thm:KLD_minimisation_Poisson}Given a multitrajectory density
$\pi\left(\cdot\right)$, the Poisson multitrajectory density $\nu\left(\cdot\right)$
that minimises the KLD $\mathrm{D}\left(\pi\left\Vert \nu\right.\right)$
is characterised by the PHD $D_{\nu}\left(\cdot\right)=D_{\pi}\left(\cdot\right)$.
\end{thm}
\phantom{}
\begin{thm}
\label{thm:KLD_minimisation_iidc}Given a multitrajectory density
$\pi\left(\cdot\right)$, the IID cluster multitrajectory density
$\nu\left(\cdot\right)$ that minimises the KLD $\mathrm{D}\left(\pi\left\Vert \nu\right.\right)$
is characterised by the PHD $D_{\nu}\left(\cdot\right)=D_{\pi}\left(\cdot\right)$
and the cardinality distribution $\rho_{\nu}\left(\cdot\right)=\rho_{\pi}\left(\cdot\right)$.
\end{thm}
Theorem \ref{thm:KLD_minimisation_Poisson} is proved in Appendix
A in \cite{Angel18_c}. The analogous theorem for sets of targets
was proved in \cite{Mahler03}. Theorem \ref{thm:KLD_minimisation_iidc}
is proved in Appendix \ref{sec:Appendix_KLD} in the supplementary
material. The analogous theorem for sets of targets was proved in
\cite{Williams15,Angel15_d}. It should be noted that, as a Poisson
RFS is a special type of IID cluster RFS, the best fitting IID cluster
RFS always has a lower or equal KLD than the best fitting Poisson
RFS.

\subsection{Inference only on alive trajectories\label{subsec:Drawback-with-cardinality}}

In this section, we explain why the TPHD and TCPHD filters are mainly
useful to approximate the posterior multitrajectory density over alive
trajectories, but not the posterior over all trajectories, which also
include dead trajectories.  This serves as a motivation to present
the TPHD and TCPHD filters for tracking only the alive trajectories
in the next sections.

Let us first explain why the TPHD filter, which considers a Poisson
approximation, is only useful for the alive trajectories \cite[Sec. V.B]{Angel18_c},
though it was derived in \cite{Angel18_c} for dead and alive trajectories.
 In the prediction step, the part of the PHD that represents a trajectory
that dies at the current time step is multiplied by the probability
of death (one minus the probability of survival) \cite[Thm. 5]{Angel18_c},
which is usually low. As time goes on, the part of the PHD that represents
dead trajectories never changes. As a result, even if a trajectory
exists with a very high probability at some point in time, once it
dies, the TPHD filter over all trajectories indicates that it existed
with a very low probability. Therefore, the TPHD does not contain
accurate information about dead trajectories, though it does contain
useful information about alive trajectories. 

In the following, we argue with an example why the TCPHD filter, which
considers an IID cluster approximation, should only consider alive
trajectories, as the TPHD filter. 
\begin{example}
Let us consider that the posterior $\pi^{k}\left(\cdot\right)$ over
the set of trajectories at time $k$ has $m$ trajectories with probability
1 so $\rho_{\pi^{k}}\left(m\right)=1$. In addition, $\pi^{k}\left(\cdot\right)$
indicates that there are $m_{d}$ dead trajectories with independent
(single trajectory) densities $\breve{d}_{1}\left(\cdot\right),...,\breve{d}_{m_{d}}\left(\cdot\right),$
and $m_{a}$ alive trajectories with independent densities $\breve{a}_{1}\left(\cdot\right),...,\breve{a}_{m_{a}}\left(\cdot\right)$,
where $m_{d}+m_{a}=m$. Note that we can obtain this kind of true
posterior, without TPHD/TCPHD approximations, if the probability of
detection is one, there is no clutter, targets are born independently
and they are far from each other at all time steps. In Appendix \ref{sec:IIDcluster_current_time}
(see supplementary material), we compute the best IID cluster density
approximation $\nu^{k}\left(\cdot\right)$ to $\pi^{k}\left(\cdot\right)$
using Theorem \ref{thm:KLD_minimisation_iidc} and show that the cardinality
distribution of the alive targets in $\nu^{k}\left(\cdot\right)$
is 
\begin{align}
\rho_{a}\left(n\right) & =\left(\begin{array}{c}
m\\
n
\end{array}\right)\left(\frac{m_{a}}{m}\right)^{n}\left(1-\frac{m_{a}}{m}\right)^{m-n},\label{eq:cardinality_alive_IID_example}
\end{align}
where $n\in\left\{ 0,1,...,m\right\} $. As the filtering recursion
continues, the total number $m$ of trajectories can only increase.
On the contrary, $m_{a}$ does not necessarily increase so after a
sufficiently long time $\frac{m_{a}}{m}$ may become very small. Then,
using the Poisson limit theorem, the cardinality distribution of the
alive targets can be approximated as Poisson with parameter $m_{a}$
\cite{Papoulis_book02}. Therefore, even in this simple example in
which the cardinality of the alive targets is known, the best IID
cluster approximation of the whole trajectory posterior approximates
the cardinality of the alive targets as a Poisson distribution. $\oblong$ 
\end{example}
The conclusion of the previous example is that, in the long run, an
IID cluster RFS is not necessarily better than a Poisson RFS, both
considered over all trajectories, to approximate the cardinality distribution
of the alive targets. In most applications, the cardinality of the
alive trajectories is considerably more important than the cardinality
of the total number of trajectories. In this paper, we therefore focus
on an IID cluster approximation of the alive trajectories to develop
the TCPHD filter. This implies that the TCPHD filter has an arbitrary
cardinality distribution for the alive targets, as the CPHD filter.

\section{Trajectory PHD filter\label{sec:Trajectory-PHD-filter}}

In this section, we derive the TPHD filter for tracking the alive
targets. The TPHD propagates the multitrajectory density of a Poisson
RFS through the filtering recursion. In the update step, the TPHD
filter uses Bayes' rule followed by a KLD minimisation to approximate
the posterior as Poisson, see Figure \ref{fig:TPHD-filter-diagram.}.
In Section \ref{subsec:Bayesian-recursion}, we present the Bayesian
filtering recursion for sets of trajectories. The prediction and update
steps of the TPHD filter are given in Sections \ref{subsec:TPHD-Prediction}
and \ref{subsec:TPHD-Update}, respectively. 

\subsection{Bayesian filtering recursion\label{subsec:Bayesian-recursion}}

The posterior multitrajectory density $\pi^{k}\left(\cdot\right)$
at time $k$, which denotes the density of set of trajectories present
at time $k$ given all measurements up to time $k$, is calculated
via the prediction and update steps:
\begin{align}
\omega^{k}\left(\mathbf{X}\right) & =\int f\left(\mathbf{X}\left|\mathbf{Y}\right.\right)\pi^{k-1}\left(\mathbf{Y}\right)\delta\mathbf{Y}\label{eq:prediction_trajectories}\\
\pi^{k}\left(\mathbf{X}\right) & =\frac{\ell^{k}\left(\mathbf{z}^{k}|\tau^{k}\left(\mathbf{X}\right)\right)\omega^{k}\left(\mathbf{X}\right)}{\ell^{k}\left(\mathbf{z}^{k}\right)}\label{eq:update_trajectories}
\end{align}
where $f\left(\cdot\left|\cdot\right.\right)$ is the transition density,
$\omega^{k}\left(\cdot\right)$ is the predicted density at time $k$,
$\mathbf{z}^{k}$ is the set of measurements at time $k$, $\ell^{k}\left(\cdot|\tau^{k}\left(\mathbf{X}\right)\right)$
is the density of the measurements given the current RFS of targets
and 
\begin{align*}
\ell^{k}\left(\mathbf{z}^{k}\right) & =\int\ell^{k}\left(\mathbf{z}^{k}|\tau^{k}\left(\mathbf{X}\right)\right)\omega^{k}\left(\mathbf{X}\right)\delta\mathbf{X}
\end{align*}
is the density of the measurements given the predicted density $\omega^{k}\left(\cdot\right)$.
The predicted density at time $k$ is the density of the set of trajectories
present at time step $k$ given the measurements up to time step $k-1$.
As we only take into account the present trajectories, the only term
that changes in (\ref{eq:prediction_trajectories})-(\ref{eq:update_trajectories})
with respect to considering all trajectories is $f\left(\cdot\left|\cdot\right.\right)$,
see \cite[Sec. IV.A]{Granstrom18} for a detailed explanation. The
description of these models will be given in Sections \ref{subsec:TPHD-Prediction}
and \ref{subsec:TPHD-Update}. 

\subsection{Prediction\label{subsec:TPHD-Prediction}}

We make the following assumptions in the prediction step: 
\begin{itemize}
\item P1 Given the current set ${\bf x}$ of targets, each target $x\in{\bf x}$
survives with probability $p_{S}\left(x\right)$ and moves to a new
state with a transition density $g\left(\cdot\left|x\right.\right)$,
or dies with probability $1-p_{S}\left(x\right)$.
\item P2 The multitarget state at the next time step is the union of the
surviving targets and new targets, which are born independently with
a Poisson multitarget density $\beta_{\tau}\left(\cdot\right)$.
\item P3 The multitrajectory density $\pi^{k-1}\left(\cdot\right)$ of the
trajectories present at time $k-1$ represents a Poisson RFS.
\end{itemize}
Note that we use subindex $\tau$ in densities on RFS of targets,
as in $\beta_{\tau}\left(\cdot\right)$. Let $\mathbb{N}_{k}=\left\{ 1,...,k\right\} $.
Then, the relation between predicted PHD at time $k$ and the PHD
of the posterior at time $k-1$ is given by the following theorem.
\begin{thm}[TPHD filter prediction]
\label{thm:PHD_prediction}Under Assumptions P1-P3, the predicted
PHD $D_{\omega^{k}}\left(\cdot\right)$ of the trajectories present
at time $k$ is
\begin{align}
D_{\omega^{k}}\left(X\right) & =D_{\xi^{k}}\left(X\right)+D_{\beta^{k}}\left(X\right)\label{eq:predicted_PHD}
\end{align}
where
\begin{align*}
D_{\beta^{k}}\left(t,x^{1:i}\right) & =D_{\beta_{\tau}}\left(x^{1}\right)1_{\left\{ k\right\} }\left(t\right)\\
D_{\xi^{k}}\left(t,x^{1:i}\right) & =p_{S}\left(x^{i-1}\right)g\left(x^{i}\left|x^{i-1}\right.\right)\\
 & \quad\times D_{\pi_{k-1}}\left(t,x^{1:i-1}\right)1_{\mathbb{N}_{k-1}}\left(t\right)
\end{align*}
if $t+i-1=k$ or zero otherwise. 
\end{thm}
This theorem is proved in \cite{Angel18_c} for a more general case
in which dead trajectories are considered. As mentioned in Section
\ref{subsec:Drawback-with-cardinality}, in this paper, we only present
the results for alive trajectories, as the results are mainly useful
in this case. The predicted PHD is the sum of the PHD $D_{\beta^{k}}\left(\cdot\right)$
of the trajectories born at time step $k$ and the PHD $D_{\xi^{k}}\left(\cdot\right)$
of the surviving trajectories. The end time of trajectory $\left(t,x^{1:i}\right)$
is $t+i-1$ so $D_{\omega^{k}}\left(t,x^{1:i}\right)$ is zero if
$t+i-1\neq k$. For the surviving trajectories, we multiply the PHD
by the transition density and the survival probability. Note that
the provided PHD characterises the Poisson RFS that represents the
predicted density.

\subsection{Update\label{subsec:TPHD-Update}}

We make the following assumptions in the update step:
\begin{itemize}
\item U1 For a given multi-target state $\mathbf{x}$ at time $k$, each
target state $x\in\mathbf{x}$ is either detected with probability
$p_{D}\left(x\right)$ and generates one measurement with density
$l\left(\cdot|x\right)$, or missed with probability $1-p_{D}\left(x\right)$. 
\item U2 The measurement $\mathbf{z}^{k}$ is the union of the target-generated
measurements and Poisson clutter with density $c\left(\cdot\right)$. 
\item U3 The multitrajectory density $\omega{}^{k}\left(\cdot\right)$ represents
a Poisson RFS.
\end{itemize}
Let $\Xi_{n,n_{z}}$ denote the set that contains all the vectors
$\sigma=\left(\sigma_{1},...,\sigma_{n}\right)$ that indicate associations
of $n_{z}$ measurements to $n$ targets, which can be either detected
or undetected. If $\sigma\in\Xi_{n,n_{z}}$, $\sigma_{i}=j\in\left\{ 1,...,n_{z}\right\} $
indicates that measurement $j$ is associated with target $i$ and
$\sigma_{i}=0$ indicates that target $i$ has not been detected.
Under Assumptions U1 and U2, which define the standard measurement
model, the density of the measurement given the state is \cite[Eq. (7.21)]{Mahler_book14}
\begin{align}
 & \ell^{k}\left(\left\{ z_{1},...,z_{n_{z}}\right\} \left|\left\{ x_{1},...,x_{n}\right\} \right.\right)\nonumber \\
 & \quad=e^{-\lambda_{c}}\left[\prod_{i=1}^{n_{z}}\lambda_{c}\breve{c}\left(z_{i}\right)\right]\left[\prod_{i=1}^{n}\left(1-p_{D}\left(x_{i}\right)\right)\right]\nonumber \\
 & \qquad\times\sum_{\sigma\in\Xi_{n,n_{z}}}\prod_{i:\sigma_{i}>0}\frac{p_{D}\left(x_{i}\right)l\left(z_{\sigma_{i}}|x_{i}\right)}{\left(1-p_{D}\left(x_{i}\right)\right)\lambda_{c}\breve{c}\left(z_{\sigma_{i}}\right)}.\label{eq:PDF_measurement_full_PHD}
\end{align}
where $\lambda_{c}$ and $\breve{c}\left(\cdot\right)$ characterise
$c\left(\cdot\right)$, see (\ref{eq:Poisson_prior}). 

Let $L_{\mathbf{z}^{k}}\left(\cdot\right)$ denote the PHD filter
pseudolikelihood function, which is given by \cite[Sec. 8.4.3]{Mahler_book14}
\begin{align*}
L_{\mathbf{z}^{k}}\left(x\right) & =1-p_{D}\left(x\right)+p_{D}\left(x\right)\\
 & \quad\times\sum_{z\in\mathbf{z}^{k}}\frac{l\left(z|x\right)}{\lambda_{c}\breve{c}\left(z\right)+\int p_{D}\left(y\right)l\left(z|y\right)D_{\omega_{\tau}^{k}}\left(y\right)dy}
\end{align*}
with $D_{\omega_{\tau}^{k}}\left(\cdot\right)$ representing the PHD
of the targets at time $k$ of density $\omega^{k}\left(\cdot\right)$,
which is given by \cite{Angel18_c} 
\begin{align}
D_{\omega_{\tau}^{k}}\left(y\right) & =\sum_{t=1}^{k}\int D_{\omega^{k}}\left(t,x^{1:k-t},y\right)dx^{1:k-t}.\label{eq:PHD_targets_prior}
\end{align}
Then, the TPHD filter update step is given by the following theorem:
\begin{thm}[TPHD filter update]
\label{thm:PHD_update}Under Assumptions U1-U3, the updated PHD $D_{\pi^{k}}\left(\cdot\right)$
at time $k$ is
\begin{align*}
D_{\pi^{k}}\left(t,x^{1:i}\right) & =D_{\omega^{k}}\left(t,x^{1:i}\right)L_{\mathbf{z}^{k}}\left(x^{i}\right)
\end{align*}
if $t+i-1=k$ or zero, otherwise.
\end{thm}
This theorem is proved in \cite{Angel18_c} for a more general case
in which dead trajectories are included. It should be noted that Bayes'
update (\ref{eq:update_trajectories}) uses a likelihood (\ref{eq:PDF_measurement_full_PHD})
that involves a summation over all target-to-measurement associations
in the multitarget space. In contrast, the TPHD filter update is similar
to the PHD filter update in the sense that it uses a pseudolikelihood
function $L_{\mathbf{z}^{k}}\left(\cdot\right)$ which is defined
on the single target space and only involves associations between
a single target and the measurements.

\section{Trajectory CPHD filter\label{sec:Trajectory-CPHD-filter}}

In this section we present the trajectory CPHD (TCPHD) filter for
tracking the alive targets. The TCPHD propagates the multitrajectory
density of an IID cluster RFS through the filtering recursion. In
the prediction and update steps, the TCPHD filter makes use of a KLD
minimisation to obtain an IID cluster approximation, see Figure \ref{fig:TCPHD-filter-diagram.}.

Prior to deriving the TCPHD filter, we provide some notation. Given
two sequences $a\left(n\right)$ and $b\left(n\right)$, $n\in\mathbb{N}\cup\left\{ 0\right\} $,
we denote
\begin{align*}
\left\langle a,b\right\rangle  & =\sum_{n=0}^{\infty}a\left(n\right)b\left(n\right).
\end{align*}
Given a set $\mathbf{z}$, the elementary symmetric function of order
$j$ is \cite{Vo07}
\begin{align}
e_{j}\left(\mathbf{z}\right) & =\sum_{\mathbf{s}\subseteq\mathbf{z},\left|S\right|=j}\left(\prod_{\zeta\in\mathbf{s}}\zeta\right)\label{eq:elemmentary_symmetric_function}
\end{align}
with $e_{0}\left(Z\right)=1$ by convention. We also use $\setminus$
to denote set subtraction.

\subsection{Prediction\label{subsec:TCPHD-Prediction}}

The TCPHD filter prediction is obtained under Assumptions P1 and the
additional assumptions
\begin{itemize}
\item P4 The multitarget state at the next time step is the union of the
surviving targets and new targets, which are born independently with
an IID cluster multitarget density $\beta_{\tau}\left(\cdot\right)$.
\item P5 The multitrajectory density $\pi^{k-1}\left(\cdot\right)$ represents
an IID cluster RFS. 
\end{itemize}
Under Assumption P4, the set of new born trajectories at time $k$
has cardinality $\rho_{\beta^{k}}\left(\cdot\right)=\rho_{\beta_{\tau}}\left(\cdot\right)$
and PHD 
\begin{align*}
D_{\beta^{k}}\left(t,x^{1:i}\right) & =\begin{cases}
D_{\beta_{\tau}}\left(x^{1}\right) & t=k,\,i=1\\
0 & \mathrm{otherwise}.
\end{cases}
\end{align*}

The TCPHD filter prediction consists of applying the usual prediction
step plus a KLD minimisation, which is performed by calculating the
cardinality distribution and PHD of the predicted density, see Figure
\ref{fig:TCPHD-filter-diagram.}. The result is given in the following
theorem. 
\begin{thm}[TCPHD filter prediction]
\label{thm:TCPHD_prediction} Under Assumptions P1, P4 and P5, the
PHD of the predicted density is the same as in the PHD filter, see
Theorem \ref{thm:PHD_prediction}. The cardinality distribution of
the predicted density is
\begin{align}
\rho_{\omega^{k}}\left(m\right) & =\sum_{j=0}^{m}\rho_{\beta^{k}}\left(m-j\right)\sum_{n=j}^{\infty}\left(\begin{array}{c}
n\\
j
\end{array}\right)\rho_{\pi^{k-1}}\left(n\right)\nonumber \\
 & \quad\times\frac{\left[\int\left(1-p_{S}(x)\right)D_{\pi_{\tau}^{k-1}}(x)dx\right]^{n-j}}{\left[\int D_{\pi_{\tau}^{k-1}}(x)dx\right]^{n}}\nonumber \\
 & \quad\times\left[\int p_{S}\left(x\right)D_{\pi_{\tau}^{k-1}}(x)dx\right]^{j}\label{eq:prediction_cardinality_CPHD}
\end{align}
where $D_{\pi_{\tau}^{k-1}}\left(\cdot\right)$ is the PHD of the
targets at time $k-1$ according to $\pi^{k-1}\left(\cdot\right)$,
which is calculated as in (\ref{eq:PHD_targets_prior}).
\end{thm}
Theorem \ref{thm:TCPHD_prediction} is proved in Appendix \ref{sec:Appendix_TCPHD_prediction}
(see supplementary material). The TCPHD filter prediction updates
the cardinality distribution as the CPHD filter. The TCPHD does not
integrate out past states of the trajectories in the PHD to keep trajectory
information, while the CPHD filter does.

\subsection{Update\label{subsec:TCPHD-Update}}

The TCPHD filter update is derived under Assumption U1 and the additional
assumptions
\begin{itemize}
\item U4 The measurement $\mathbf{z}^{k}$ is the union of the target-generated
measurements and IID cluster clutter with density $c\left(\cdot\right)$. 
\item U5 The multitrajectory density $\omega{}^{k}\left(\cdot\right)$ represents
an IID cluster RFS.
\end{itemize}
As indicated in Figure \ref{fig:TCPHD-filter-diagram.}, the TCPHD
update consists of applying Bayes' rule, see (\ref{eq:update_trajectories}),
followed by a KLD minimisation, which is performed as indicated by
Theorem \ref{thm:KLD_minimisation_iidc}. We first consider the distribution
of the present targets at the current time. Under Assumption U5, it
is direct to obtain that the distribution of the targets present at
time $k$ is also an IID cluster with cardinality distribution $\rho_{\omega^{k}}\left(\cdot\right)$
and PHD (\ref{eq:PHD_targets_prior}). The resulting TCPHD filter
update is given in the following theorem.
\begin{thm}[TCPHD filter update]
\label{thm:TCPHD_update}Under Assumptions U1, U4 and U5, the cardinality
distribution and the PHD of the posterior at time $k$ are
\begin{align}
\rho_{\pi^{k}}\left(n\right) & =\frac{\Upsilon^{0}\left[D_{\omega_{\tau}^{k}},\mathbf{z}^{k}\right]\left(n\right)\rho_{\omega^{k}}\left(n\right)}{\left\langle \Upsilon^{0}\left[D_{\omega_{\tau}^{k}},\mathbf{z}^{k}\right],\rho_{\omega^{k}}\right\rangle }\label{eq:cardinality_update_CPHD}\\
D_{\pi^{k}}\left(t,x^{1:i}\right) & =\frac{\left\langle \Upsilon^{1}\left[D_{\omega_{\tau}^{k}},\mathbf{z}^{k}\right],\rho_{\omega^{k}}\right\rangle }{\left\langle \Upsilon^{0}\left[D_{\omega_{\tau}^{k}},\mathbf{z}^{k}\right],\rho_{\omega^{k}}\right\rangle }\nonumber \\
 & \quad\times\left(1-p_{D}\left(x^{i}\right)\right)D_{\omega^{k}}\left(t,x^{1:i}\right)\nonumber \\
 & \quad+\sum_{z\in\mathbf{z}^{k}}\frac{\left\langle \Upsilon^{1}\left[D_{\omega_{\tau}^{k}},\mathbf{z}^{k}\setminus\left\{ z\right\} \right],\rho_{\omega^{k}}\right\rangle }{\left\langle \Upsilon^{0}\left[D_{\omega_{\tau}^{k}},\mathbf{z}^{k}\right],\rho_{\omega^{k}}\right\rangle }\nonumber \\
 & \quad\times\frac{l\left(z|x^{i}\right)}{\breve{c}\left(z\right)}p_{D}\left(x^{i}\right)D_{\omega^{k}}(t,x^{1:i})\label{eq:PHD_update_CPHD}
\end{align}
if $t+i-1=k$ or $D_{\pi^{k}}\left(t,x^{1:i}\right)=0$, otherwise,
and
\begin{align}
\Upsilon^{u}\left[D_{\omega_{\tau}^{k}},\mathbf{z}^{k}\right]\left(n\right) & =\sum_{j=0}^{\min\left(\left|\mathbf{z}^{k}\right|,n-u\right)}\left(\left|\mathbf{z}^{k}\right|-j\right)!\rho_{c}\left(\left|\mathbf{z}^{k}\right|-j\right)\nonumber \\
 & \quad\times\frac{\left[\int\left(1-p_{D}(x)\right)D_{\omega_{\tau}^{k}}(x)dx\right]^{n-\left(j+u\right)}}{\left[\int D_{\omega_{\tau}^{k}}(x)dx\right]^{n}}\nonumber \\
 & \quad\times\frac{n!}{\left(n-j-u\right)!}e_{j}\left(\Xi\left(D_{\omega_{\tau}^{k}},\mathbf{z}^{k}\right)\right)\label{eq:gamma_u_cphd}\\
\Xi\left(D_{\omega_{\tau}^{k}},\mathbf{z}^{k}\right) & =\left\{ \int\frac{l\left(z|x\right)}{\breve{c}\left(z\right)}p_{D}\left(x\right)D_{\omega_{\tau}^{k}}(x)dx:z\in\mathbf{z}^{k}\right\} .\nonumber 
\end{align}
\end{thm}
Theorem \ref{thm:TCPHD_update} is proved in Appendix \ref{sec:Appendix_TCPHD_update}
(see supplementary material). The update step of the TCPHD filter
is equivalent to the CPHD filter, with the main difference that the
updated PHD contains information about previous states of the trajectories. 

\section{Gaussian mixture implementations\label{sec:Gaussian-mixture-implementations}}

In this section, we present the Gaussian mixture implementations of
the TPHD and TCPHD filters. We use the notation 
\begin{align}
\mathcal{N}\left(t,x^{1:i};t^{k},m^{k},P^{k}\right) & =\begin{cases}
\mathcal{N}\left(x^{1:i};m^{k},P^{k}\right) & t=t^{k},\,i=i^{k}\\
0 & \mathrm{otherwise}
\end{cases}\label{eq:Trajectory_Gaussian}
\end{align}
where $i^{k}=\mathrm{dim}\left(m^{k}\right)/n_{x}$. Equation (\ref{eq:Trajectory_Gaussian})
represents a single trajectory Gaussian density with start time $t^{k}$,
duration $i^{k}$, mean $m^{k}\in\mathbb{R}^{i^{k}n_{x}}$ and covariance
matrix $P^{k}\in\mathbb{R}^{i^{k}n_{x}\times i^{k}n_{x}}$ evaluated
at $\left(t,x^{1:i}\right)$. We use $\otimes$ to indicate the Kronecker
product and $0_{m,n}$ is the $m\times n$ zero matrix. 

We make the additional assumptions
\begin{itemize}
\item A1 The probabilities $p_{S}$ and $p_{D}$ are constants. 
\item A2 $g\left(x^{i}\left|x^{i-1}\right.\right)=\mathcal{N}\left(x^{i};Fx^{i-1},Q\right)$. 
\item A3 $l\left(z|x\right)=\mathcal{N}\left(z;Hx,R\right)$.
\item A4 The PHD of the birth density $\beta^{k}\left(\cdot\right)$ is
\begin{align}
D_{\beta^{k}}\left(X\right) & =\sum_{j=1}^{J_{\beta}^{k}}w_{\beta,j}^{k}\mathcal{N}\left(X;k,m_{\beta,j}^{k},P_{\beta,j}^{k}\right)\label{eq:GMPHD-birth}
\end{align}
where $J_{\beta}^{k}\in\mathbb{N}$ is the number of components, $w_{\beta,j}^{k}$
is the weight of the $j$th component, $m_{\beta,j}^{k}\in\mathbb{R}^{n_{x}}$
its mean and $P_{\beta,j}^{k}\in\mathbb{R}^{n_{x}\times n_{x}}$ its
covariance matrix. 
\end{itemize}
It should be noted that $F\in\mathbb{R}^{n_{x}\times n_{x}}$ is the
single-target transition matrix, $Q\in\mathbb{R}^{n_{x}\times n_{x}}$
is the covariance matrix of the single-target process noise, $H\in\mathbb{R}^{n_{z}\times n_{x}}$
is the single-measurement matrix and $R\in\mathbb{R}^{n_{z}\times n_{z}}$
is the covariance matrix of the single-measurement noise. In addition,
the models provided by A1-A4 could be time varying but time is omitted
for notational convenience. In the rest of this section, we present
the Gaussian mixture implementations of the TPHD and TCPHD filters
in Sections \ref{subsec:GMTPHD-Prediction-and-update} and \ref{subsec:GMTCPHD-Prediction-and-update},
respectively. The $L$-scan versions of the filters and trajectory
estimation are addressed in Sections \ref{subsec:L-scan-GMTPHD} and
\ref{subsec:Estimation}. Finally, a discussion is provided in Section
\ref{subsec:Discussion}.

\subsection{Gaussian mixture TPHD filter\label{subsec:GMTPHD-Prediction-and-update}}

Under Assumptions A1-A4, P1-P3 and U1-U3, we can calculate the TPHD
filter in closed form giving rise to the GMTPHD filter, whose prediction
and update steps are provided in the following propositions. 
\begin{prop}[GMTPHD filter prediction]
\label{prop:GMTPHD_prediction}Assume $\pi^{k-1}\left(\cdot\right)$
has a PHD
\begin{align*}
D_{\pi^{k-1}}\left(X\right) & =\sum_{j=1}^{J^{k-1}}w_{j}^{k-1}\mathcal{N}\left(X;t_{j}^{k-1},m_{j}^{k-1},P_{j}^{k-1}\right)
\end{align*}
where $t_{j}^{k-1}+i_{j}^{k-1}-1=k-1$ with $i_{j}^{k-1}=\mathrm{dim}\left(m_{j}^{k-1}\right)/n_{x}$.
Then, the PHD of $\omega^{k}\left(\cdot\right)$ is
\begin{align}
D_{\omega^{k}}\left(X\right) & =D_{\beta^{k}}\left(X\right)+p_{S}\sum_{j=1}^{J^{k-1}}w_{j}^{k-1}\mathcal{N}\left(X;t_{j}^{k-1},m_{\omega,j}^{k},P_{\omega,j}^{k}\right)\label{eq:GMTPHD_prediction}
\end{align}
where
\begin{align*}
m_{\omega,j}^{k} & =\left[\left(m_{j}^{k-1}\right)^{T},\left(\dot{F}_{j}m_{j}^{k-1}\right)^{T}\right]^{T},\\
\dot{F}_{j} & =\left[0_{1,i_{j}^{k-1}-1},1\right]\otimes F,\\
P_{\omega,j}^{k} & =\left[\begin{array}{cc}
P_{j}^{k-1} & P_{j}^{k-1}\dot{F}_{j}^{T}\\
\dot{F}_{j}P_{j}^{k-1} & \dot{F}_{j}P_{j}^{k-1}\dot{F}_{j}^{T}+Q
\end{array}\right].
\end{align*}
\end{prop}
Proposition \ref{prop:GMTPHD_prediction} is a consequence of Theorem
\ref{thm:PHD_prediction} and conventional properties of Gaussian
densities. Compared to the GMPHD filter prediction, the main difference
is that previous states are not integrated out. 
\begin{prop}[GMTPHD filter update]
\label{prop:GMTPHD_update}Assume $\omega^{k}\left(\cdot\right)$
has a PHD
\begin{align}
D_{\omega^{k}}\left(X\right) & =\sum_{j=1}^{J_{\omega}^{k}}w_{\omega,j}^{k}\mathcal{N}\left(X;t_{\omega,j}^{k},m_{\omega,j}^{k},P_{\omega,j}^{k}\right).\label{eq:GMTPHD_predicted_PHD_update}
\end{align}
Then, the PHD of $\pi^{k}\left(\cdot\right)$ is
\begin{align}
D_{\pi^{k}}\left(X\right) & =\left(1-p_{D}\right)D_{\omega^{k}}\left(X\right)\nonumber \\
 & \,+\sum_{z\in\mathbf{z}^{k}}\sum_{j=1}^{J^{k}}w_{j}\left(z\right)\mathcal{N}\left(X;t_{\omega,j}^{k},m_{j}^{k}\left(z\right),P_{j}^{k}\right)\label{eq:GMTPHD_update}
\end{align}
where
\begin{align*}
w_{j}\left(z\right) & =\frac{p_{D}w_{\omega,j}^{k}q_{j}\left(z\right)}{\lambda_{c}\breve{c}\left(z\right)+p_{D}\sum_{l=1}^{J_{\omega}^{k}}w_{\omega,l}^{k}q_{l}\left(z\right)}\\
\overline{z}_{j} & =\dot{H}_{j}m_{\omega,j}^{k},\quad S_{j}=\dot{H}_{j}P_{\omega,j}^{k}\dot{H}_{j}^{T}+R\\
\dot{H}_{j} & =\left[0_{1,i_{\omega,j}^{k}-1},1\right]\otimes H\\
q_{j}\left(z\right) & =\mathcal{N}\left(z;\overline{z}_{j},S_{j}\right)\\
m_{j}^{k}\left(z\right) & =m_{\omega,j}^{k}+P_{\omega,j}^{k}\dot{H}^{T}S_{j}^{-1}\left(z-\overline{z}_{j}\right)\\
P_{j}^{k} & =P_{\omega,j}^{k}-P_{\omega,j}^{k}\dot{H}^{T}S_{j}^{-1}\dot{H}P_{\omega,j}^{k}.
\end{align*}
and $i_{\omega,j}^{k}=\mathrm{dim}\left(m_{\omega,j}^{k}\right)/n_{x}$
\end{prop}
Proposition \ref{prop:GMTPHD_update} is a consequence of Theorem
\ref{thm:PHD_update} and the Kalman filter update equations \cite{Sarkka_book13}.
Also, the GMTPHD filter update is similar to the GMPHD filter update.
The main differences is that the GMTPHD updates the whole trajectories.

\subsection{Gaussian mixture TCPHD filter\label{subsec:GMTCPHD-Prediction-and-update}}

The GMTCPHD filtering recursion requires Assumptions A1-A4, P1, P4,
P5, U1, U4 and U5, and is given by the following propositions. 
\begin{prop}[GMTCPHD filter prediction]
\label{prop:GMTCPHD_prediction}Assume the posterior $\pi^{k-1}\left(\cdot\right)$
has a cardinality distribution $\rho_{\pi^{k-1}}\left(\cdot\right)$
and a PHD
\begin{align}
D_{\pi^{k-1}}\left(X\right) & =\sum_{j=1}^{J^{k-1}}w_{j}^{k-1}\mathcal{N}\left(X;t_{j}^{k-1},m_{j}^{k-1},P_{j}^{k-1}\right).\label{eq:GMTCPHD-posterior}
\end{align}
Then, the TCPHD filter prediction yields
\begin{align*}
\rho_{\omega^{k}}\left(m\right) & =\sum_{j=0}^{m}\rho_{\beta^{k}}\left(m-j\right)\\
 & \quad\times\sum_{n=j}^{\infty}\left(\begin{array}{c}
n\\
j
\end{array}\right)\rho_{\pi^{k-1}}\left(n\right)\left(1-p_{S}\right)^{n-j}p_{S}^{j}\\
D_{\omega^{k}}\left(X\right) & =D_{\beta^{k}}\left(X\right)+p_{S}\sum_{j=1}^{J^{k-1}}w_{j}^{k-1}\mathcal{N}\left(X;t_{j}^{k-1},m_{\omega,j}^{k},P_{\omega,j}^{k}\right)
\end{align*}
where $m_{\omega,j}^{k}$ and $P_{\omega,j}^{k}$ are given in Proposition
\ref{prop:GMTPHD_prediction}. 
\end{prop}
Proposition \ref{prop:GMTCPHD_prediction} is a consequence of Theorem
\ref{thm:TCPHD_prediction}. The main difference with the GMCPHD filter
is that the GMTCPHD filter does not integrate out previous states
in the PHD. It should also be noted that the GMTPHD and the GMTCPHD
propagate the PHD in the same way. The difference is that the GMTCPHD
also considers the cardinality of $\omega^{k}\left(\cdot\right)$.
\begin{prop}[GMTCPHD filter update]
\label{prop:GMTCPHD_update}Assume the prior $\omega^{k}\left(\cdot\right)$
has a cardinality distribution $\rho_{\omega^{k}}\left(\cdot\right)$
and a PHD 
\begin{align}
D_{\omega^{k}}\left(X\right) & =\sum_{j=1}^{J_{\omega}^{k}}w_{\omega,j}^{k}\mathcal{N}\left(X;t_{\omega,j}^{k},m_{\omega,j}^{k},P_{\omega,j}^{k}\right).\label{eq:GMCPHD_prior}
\end{align}
Then, the TCPHD filter update yields 
\begin{align*}
\rho_{\pi^{k}}\left(n\right) & =\frac{\Psi^{0}\left[w_{\omega}^{k},\mathbf{z}^{k}\right]\left(n\right)\rho_{\omega^{k}}\left(n\right)}{\left\langle \Psi^{0}\left[w_{\omega}^{k},\mathbf{z}^{k}\right],\rho_{\omega^{k}}\right\rangle }\\
D_{\pi^{k}}\left(X\right) & =\frac{\left\langle \Psi^{1}\left[w_{\omega}^{k},\mathbf{z}^{k}\right],\rho_{\omega^{k}}\right\rangle }{\left\langle \Psi^{0}\left[w_{\omega}^{k},\mathbf{z}^{k}\right],\rho_{\omega^{k}}\right\rangle }\left(1-p_{D}\right)D_{\omega^{k}}\left(X\right)\\
 & \quad+\sum_{z\in\mathbf{z}^{k}}\sum_{j=1}^{J_{\omega}^{k}}w_{j}\left(z\right)\mathcal{N}\left(X;t_{\omega,j}^{k},m_{j}^{k}\left(z\right),P_{j}^{k}\right)
\end{align*}
where
\begin{align*}
\Psi^{u}\left[w_{\omega}^{k},\mathbf{z}^{k}\right]\left(n\right) & =\sum_{j=0}^{\min\left(\left|\mathbf{z}^{k}\right|,n-u\right)}\left(\left|\mathbf{z}^{k}\right|-j\right)!\rho_{c}\left(\left|\mathbf{z}^{k}\right|-j\right)\\
 & \quad\times\frac{\left(1-p_{D}\right)^{n-\left(j+u\right)}}{\left\langle 1,w_{\omega}^{k}\right\rangle ^{j+u}}\frac{n!}{\left(n-j-u\right)!}\\
 & \quad\times e_{j}\left(\Lambda\left(w_{\omega}^{k},\mathbf{z}^{k}\right)\right)\\
\Lambda\left(w_{\omega}^{k},\mathbf{z}^{k}\right) & =\left\{ \frac{p_{D}}{\breve{c}\left(z\right)}\left(w_{\omega}^{k}\right)^{T}q\left(z\right):z\in\mathbf{z}^{k}\right\} \\
w_{\omega}^{k} & =\left[w_{\omega,1}^{k},...,w_{\omega,J_{\omega}^{k}}^{k}\right]^{T}\\
q\left(z\right) & =\left[q_{1}\left(z\right),...,q_{J_{\omega}^{k}}\left(z\right)\right]^{T}\\
w_{j}\left(z\right) & =\frac{p_{D}w_{\omega,j}^{k}q_{j}\left(z\right)\left\langle \Psi^{1}\left[w_{\omega}^{k},\mathbf{z}^{k}\setminus\left\{ z\right\} \right],\rho_{\omega^{k}}\right\rangle }{\breve{c}\left(z\right)\left\langle \Psi^{0}\left[w_{\omega}^{k},\mathbf{z}^{k}\right],\rho_{\omega^{k}}\right\rangle }
\end{align*}
and $q_{j}\left(z\right)$, $m_{j}^{k}\left(z\right)$ and $P_{j}^{k}$
are given in Proposition \ref{prop:GMTPHD_update}. 
\end{prop}
Proposition \ref{prop:GMTCPHD_update} is a consequence of Theorem
\ref{thm:TCPHD_update} and the Kalman filter update \cite{Sarkka_book13}.
The GMTCPHD filter update is analogous to the GMCPHD filter \cite{Vo07},
with the difference that previous states of the target trajectories
are also included. 

\subsection{$L$-scan implementations\label{subsec:L-scan-GMTPHD}}

In this section, we propose the use of pruning and absorption to limit
the number of components in the Gaussian mixture and a computationally
efficient implementation of the Gaussian mixture filters: the $L$-scan
GMTPHD and GMTCPHD filters.

The PHD of the GMTPHD/GMTCPHD filters has an increasing number of
components as time progresses and, to limit complexity, we need to
bound the number of components. We use the following techniques: pruning
with threshold $\Gamma_{p}$, setting a maximum number $J_{max}$
of components and absorption \cite{Angel18_c}. Absorption consists
of removing components of the PHD whose distribution of the current
target state is close to the distribution of the current target state
of another component with a higher weight, and adding the weights
of the removed components to the weight of the component that has
not been removed. Absorption is motivated by the fact that if two
components have a very similar distribution over the current target
state, based on a Mahalanobis distance criterion, future measurements
will affect both component weights and future states in a similar
way. Therefore, without absorption, we would have two components with
practically the same Gaussian densities for the trajectory states
corresponding to recent time steps, for which we would be repeating
the same calculations. It should also be noted that single trajectory
densities can be quite different in the past even if they are similar
for the current target state. Therefore, a direct use of merging \cite{Salmond09}
for single trajectory densities, which would use moment matching at
all time steps, can provide poor results and absorption is preferred.
The steps of the pruning and absorption algorithms for the GMTPHD/GMTCPHD
filters are given in Algorithm \ref{alg:Pruning-and-absorption},
where we use the notation $\Phi_{j}^{k}=\left(w_{j}^{k},t_{j}^{k},m_{j}^{k},P_{j}^{k}\right)$
. 

\begin{algorithm}
\caption{\label{alg:Pruning-and-absorption}Pruning and absorption for the
GMTPHD and GMTCPHD filters}

{\fontsize{9}{9}\selectfont 

\textbf{Input: }Posterior PHD parameters $\left\{ \Phi_{j}^{k}\right\} _{j=1}^{J^{k}}$,
pruning threshold $\Gamma_{p}$, absorption threshold $\Gamma_{a}$,
maximum number of terms $J_{max}$. 

\textbf{Output:} Pruned posterior PHD parameters $\left\{ \Phi_{o,j}^{k}\right\} _{j=1}^{\hat{J}^{k}}$

\begin{algorithmic}     

\State - Set $l=0$ and $I=\left\{ j\in\left\{ 1,...,J^{k}\right\} :w_{j}^{k}>\Gamma_{p}\right\} $.

\While{$I\neq\emptyset$}

\State - Set $l\leftarrow l+1$.

\State - $j=\underset{i\in I}{\arg\max}\:w_{i}^{k}$.

\State - $L=\left\{ i\in I:\left(\hat{m}_{i}^{k}-\hat{m}_{j}^{k}\right)^{T}\left(\hat{P}_{j}^{k}\right)^{-1}\left(\hat{m}_{i}^{k}-\hat{m}_{j}^{k}\right)\leq\Gamma_{a}\right\} $
with $\hat{m}_{j}^{k}\in\mathbb{R}^{n_{x}}$ and $\hat{P}_{j}^{k}\in\mathbb{R}^{n_{x}\times n_{x}}$
denoting the mean and covariance matrix of the state at the current
time step of the PHD component indexed by $j$.

\State - $\Phi_{o,l}^{k}=\Phi_{j}^{k}$ with weight $w_{o,l}^{k}=\sum_{i\in L}w_{i}^{k}$.

\State - $I\leftarrow I\setminus L$.

\EndWhile

\State - If $l>J_{max}$, only keep the $J_{max}$ components with
highest weight.

\end{algorithmic}

}
\end{algorithm}

In addition, as time progresses, the lengths of the trajectories increase
so, eventually, it is not computationally feasible to implement the
proposed filters directly. In order to address this problem, we propose
the $L$-scan implementations that propagate the joint density of
the states of the last $L$ time steps and independent densities for
the previous states for each component of the PHD. This approach has
a Kullback-Leibler divergence interpretation \cite{Angel18_c} and
is motivated by the fact that measurements at the current time step
only have a significant impact on the trajectory state estimates for
recent time steps.

The $L$-scan GMTPHD/GMTCPHD filters are implemented as the GMTPHD/GMTCPHD
with a minor modification in the prediction step, where we discard
the correlations of states that happened $L$ time steps before the
current time step. Given the predicted PHD $D_{\omega^{k}}\left(\cdot\right)$
in Gaussian mixture form, see (\ref{eq:GMTPHD_prediction}), its $L$-scan
version is given by approximating the covariance matrices $P_{\omega,j}^{k}$
as
\begin{align}
P_{\omega,j}^{k} & \approx\mathrm{diag}\left(\tilde{P}_{j}^{t_{\omega,j}^{k}},\tilde{P}_{j}^{t_{\omega,j}^{k}+1},...,\tilde{P}_{j}^{k-L},\tilde{P}_{j}^{k-L+1:k}\right)\label{eq:L_scan_cov_approx}
\end{align}
where matrix $\tilde{P}_{j}^{k-L+1:k}\in\mathbb{R}^{L\cdot n_{x}\times L\cdot n_{x}}$
represents the joint covariance of the $L$ last time instants, obtained
from $P_{\omega,j}^{k}$, and $\tilde{P}_{j}^{k}\in\mathbb{R}^{n_{x}\times n_{x}}$
represents the covariance matrix of the target state at time $k$,
obtained from $P_{\omega,j}^{k}$. Therefore, we have independent
Gaussian densities to represent the states outside the $L$-scan window,
and a joint Gaussian density for the states in the $L$-scan window,
as in \cite{Koch11}. The steps of the $L$-scan GMTPHD and GMTCPHD
filters are summarised in Algorithm \ref{alg:L-scan_algorithms}.
It should be noted that the cardinality distribution is not affected
by the choice of $L$ in both filters. 

\begin{algorithm}
\caption{\label{alg:L-scan_algorithms}$L$-scan Gaussian mixture TPHD/TCPHD
filter steps}

{\fontsize{9}{9}\selectfont

\begin{algorithmic}     

\State - Initialisation: 

\State $\quad$- For TPHD: $D_{\pi^{0}}\left(\cdot\right)=0$.

\State $\quad$- For TCPHD: $D_{\pi^{0}}\left(\cdot\right)=0$, $\rho_{\pi^{0}}\left(0\right)=1$. 

\For{ $k=1$ to \textit{final time step} }

\State - Prediction: 

\State $\quad$- For TPHD: use Proposition \ref{prop:GMTPHD_prediction}.

\State $\quad$- For TCPHD: use Proposition \ref{prop:GMTCPHD_prediction}. 

\State $\quad$- Approximate $P_{\omega,j}^{k}$ using (\ref{eq:L_scan_cov_approx}),
which discards correlations outside the $L$-scan window.

\State - Update:

\State $\quad$- For TPHD: use Proposition \ref{prop:GMTPHD_update}. 

\State $\quad$- For TCPHD: use Proposition \ref{prop:GMTCPHD_update}. 

\State - Perform pruning/absorption using Algorithm \ref{alg:Pruning-and-absorption}.

\State - Estimate the alive trajectories, see Section \ref{subsec:Estimation}. 

\EndFor

\end{algorithmic}

}
\end{algorithm}

\subsection{Estimation\label{subsec:Estimation}}

In this section, we adapt two commonly used estimators of the GMPHD
and GMCPHD filters to the GMTPHD and GMTCPHD filters. We have observed
via simulations that better performance is obtained if these estimators
are applied after the pruning/absorption step, as indicated in Algorithm
\ref{alg:L-scan_algorithms}.

\subsubsection{GMTPHD\label{subsec:GMTPHD-Estimation}}

We adapt the estimator for the GMPHD filter described in \cite[Sec. 9.5.4.4]{Mahler_book14}
for sets of trajectories. First, the number of trajectories is estimated
as
\begin{align}
\hat{N}^{k} & =\mathrm{round}\left(\sum_{j=1}^{J^{k}}w_{j}^{k}\right).\label{eq:estimated_trajectories}
\end{align}
Then, the estimated set of trajectories corresponds to $\left\{ \left(t_{l_{1}}^{k},m_{l_{1}}^{k}\right),...,\left(t_{l_{\hat{N}^{k}}}^{k},m_{l_{\hat{N}^{k}}}^{k}\right)\right\} $
where $\left\{ l_{1},...,l_{\hat{N}^{k}}\right\} $ are the indices
of the PHD components with highest weights. 

\subsubsection{GMTCPHD\label{subsec:GMTCPHD-Estimation}}

We adapt the estimator for the GMCPHD filter described in \cite[Sec. 9.5.5.4]{Mahler_book14}
for sets of trajectories. The estimated cardinality at time step $k$
is obtained as
\begin{align}
\hat{N}^{k} & =\underset{n\in\mathbb{N}\cup\left\{ 0\right\} }{\arg\max}\,\rho_{\pi^{k}}\left(n\right).\label{eq:cardinality_estimate}
\end{align}
Then, the estimated set of trajectories are given by $\left\{ \left(t_{l_{1}}^{k},m_{l_{1}}^{k}\right),...,\left(t_{l_{\hat{N}^{k}}}^{k},m_{l_{\hat{N}^{k}}}^{k}\right)\right\} $
where $\left\{ l_{1},...,l_{\hat{N}^{k}}\right\} $ are the indices
of the PHD components with highest weights.

\subsection{Discussion\label{subsec:Discussion}}

In this section, we discuss some of the prominent aspects of the proposed
filters. The TPHD/TCPHD filters have a similar structure as the PHD/CPHD
filters, with the additional benefit of providing trajectory estimates
for the alive targets. That is, at each time step, these filters can
estimate the trajectories of each of the targets, including its time
of birth. The main difference between the trajectory PHD/CPHD filters
and their target counterparts is that the trajectory filters do not
integrate out previous states of the trajectories. 

The $L$-scan GMTPHD/GMTCPHD filters are computationally efficient
implementations, which allow the propagation of the posterior for
long time sequences. In fact, the 1-scan versions ($L=1$) of the
GMTPHD/GMTCPHD filters perform the same computations as the GMPHD/GMCPHD,
the only difference being that the trajectory versions store the mean
and covariance of the trajectory state at each time instant for each
component of the PHD. 

It should be noted that the computations of the cardinality probability
mass functions and elementary symmetric functions are the same as
in the GMCPHD filter. To calculate the elementary symmetric functions,
we use the recursive formula in \cite[Eq. (2.3)]{Oruc04}. We would
also like to note that we have presented the filters considering linear
and Gaussian models for ease of exposition. Nevertheless, we can still
apply the GMTPHD and GMTCPHD filters for nonlinear measurement and
dynamic models by first linearising the system and then applying the
prediction and update steps with the linearised model. This is the
usual procedure in nonlinear Gaussian filtering, for example, as in
the extended Kalman filter, the unscented Kalman filter or the iterated
posterior linearisation filter \cite{Sarkka_book13,Angel15_c}. 

\section{Simulations\label{sec:Simulations}}

We proceed to assess the performance of the two proposed filters in
comparison with the previous track building procedures for PHD/CPHD
filters based on tagging each PHD component \cite{Panta09}. In this
track building approach, we estimate the number $\hat{n}_{k}$ of
targets as indicated in Section \ref{subsec:Estimation} and take
the $\hat{n}_{k}$ highest components of the PHD with different tags.
These estimates are then appended to the estimated trajectories with
the same tag at the previous time step. We refer to these algorithms
as tagged PHD/CPHD filters. All units of the quantities in this section
are given in the international system.

We consider a target state $x=\left[p_{x},\dot{p}_{x},p_{y},\dot{p}_{y}\right]^{T}$,
which contains position and velocity. The parameters of the single-target
dynamic process are 
\begin{align*}
F=I_{2}\otimes\left(\begin{array}{cc}
1 & \tau\\
0 & 1
\end{array}\right),\quad Q=qI_{2}\otimes\left(\begin{array}{cc}
\tau^{3}/3 & \tau^{2}/2\\
\tau^{2}/2 & \tau
\end{array}\right)
\end{align*}
where $\tau=0.5$ is the sampling time and $q=3.24$ is a parameter.
We also set $p_{S}=0.99$. The parameters of the measurement model
are 
\begin{align*}
H=\left(\begin{array}{cccc}
1 & 0 & 0 & 0\\
0 & 0 & 1 & 0
\end{array}\right),\quad R=\sigma^{2}I_{2},
\end{align*}
where $\sigma^{2}=4$, and $p_{D}=0.9$. The clutter intensity is
$D_{c}\left(z\right)=\lambda_{c}\cdot u_{A}\left(z\right)$ where
$u_{A}\left(z\right)$ is a uniform density in region $A=\left[0,2000\right]\times\left[0,2000\right]$
and $\lambda_{c}=50$ is the average number of clutter measurements
per scan. The birth process is Poisson with a PHD that is represented
by a Gaussian mixture with parameters: $J_{\beta}^{k}=3$, $w_{\beta,j}^{k}=0.1$,
$P_{\beta,j}^{k}=\mathrm{diag}\left(\left[225,100,225,100\right]\right)$
for $j\in\left\{ 1,2,3\right\} $, $m_{\beta,1}^{k}=\left[85,0,140,0\right]^{T}$,
$m_{\beta,2}^{k}=\left[-5,0,220,0\right]^{T}$ and $m_{\beta,3}^{k}=\left[7,0,50,0\right]^{T}$. 

\begin{figure}
\begin{centering}
\includegraphics[scale=0.37]{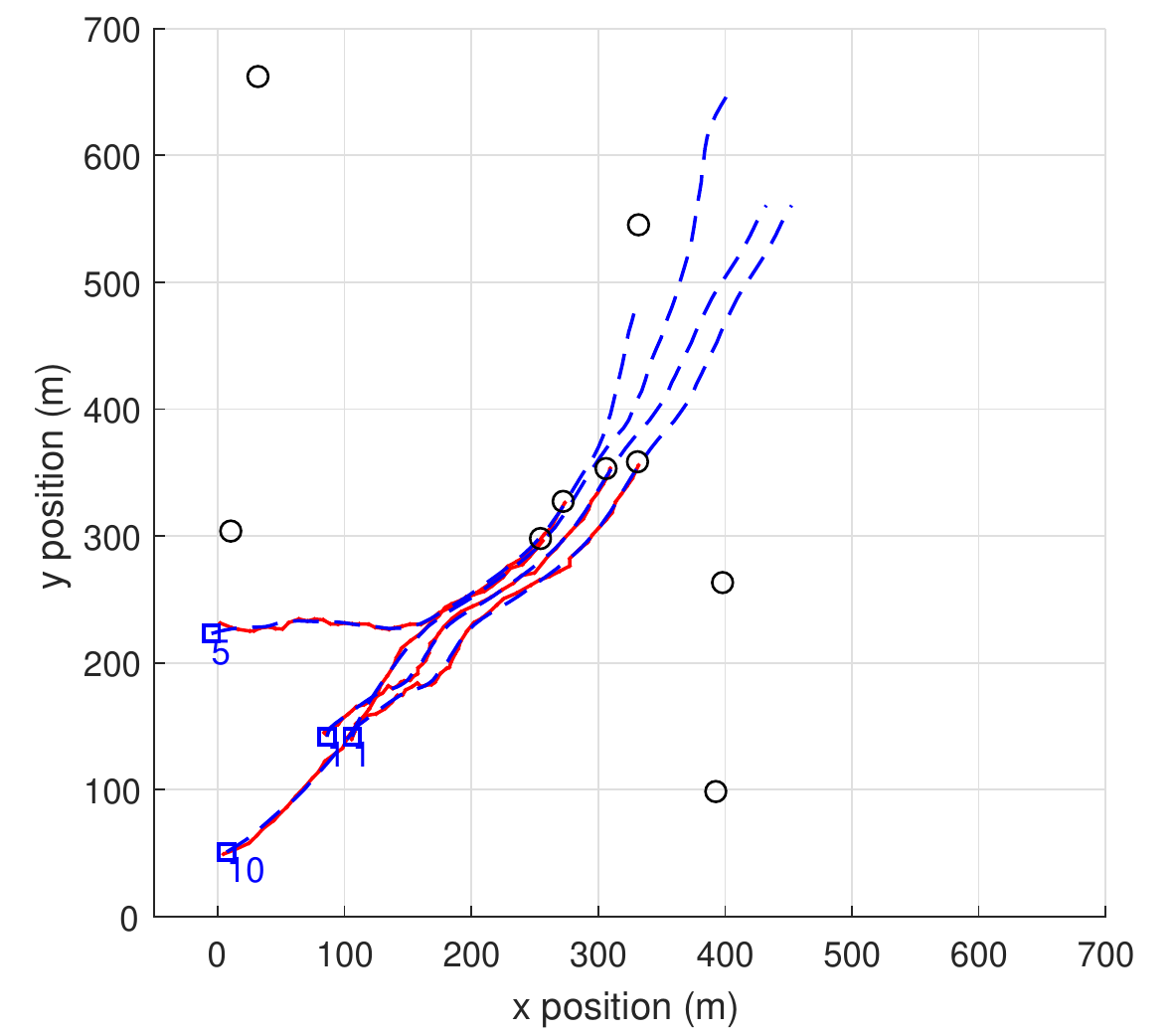}\includegraphics[scale=0.37]{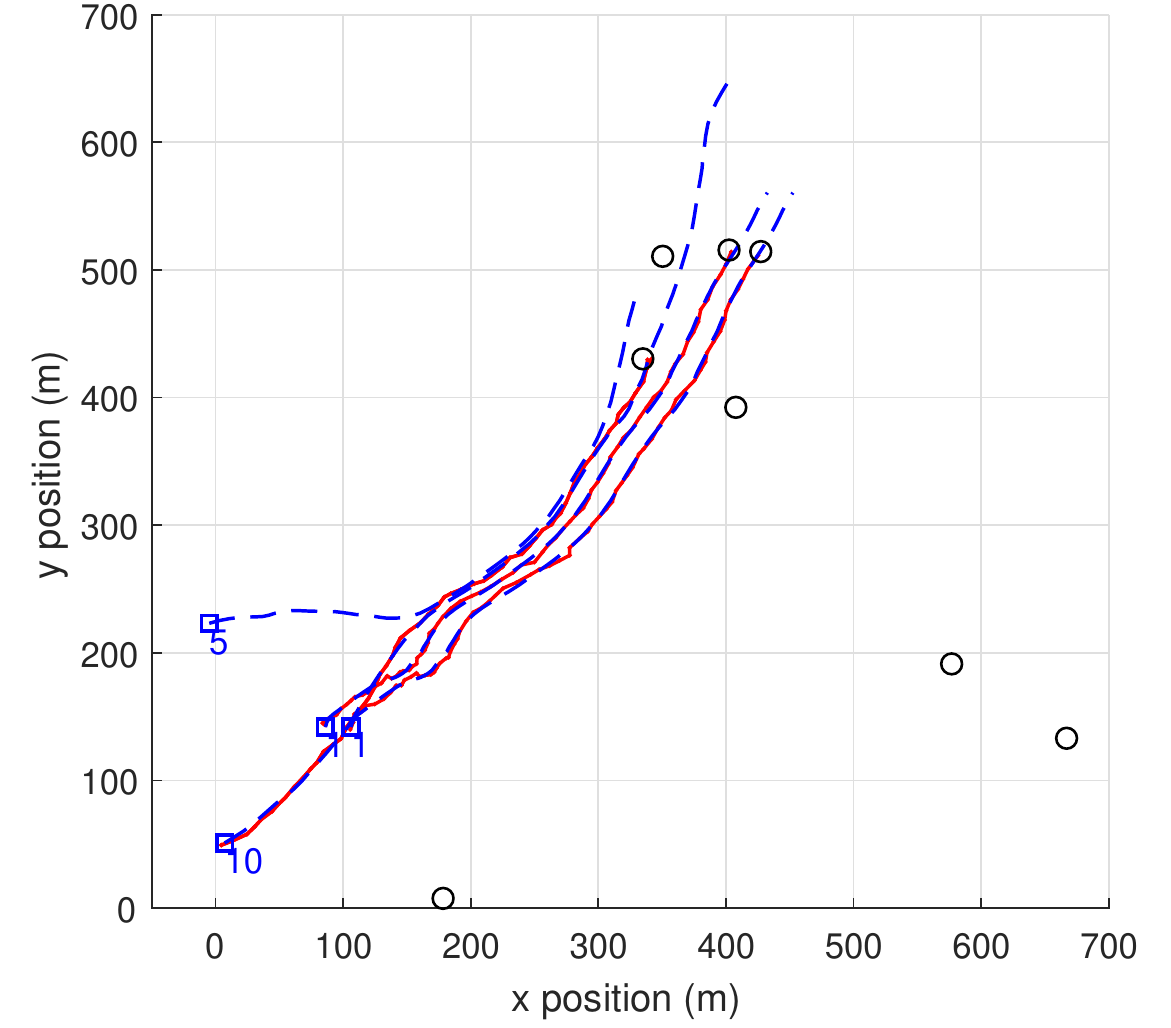}
\par\end{centering}
\begin{centering}
\includegraphics[scale=0.37]{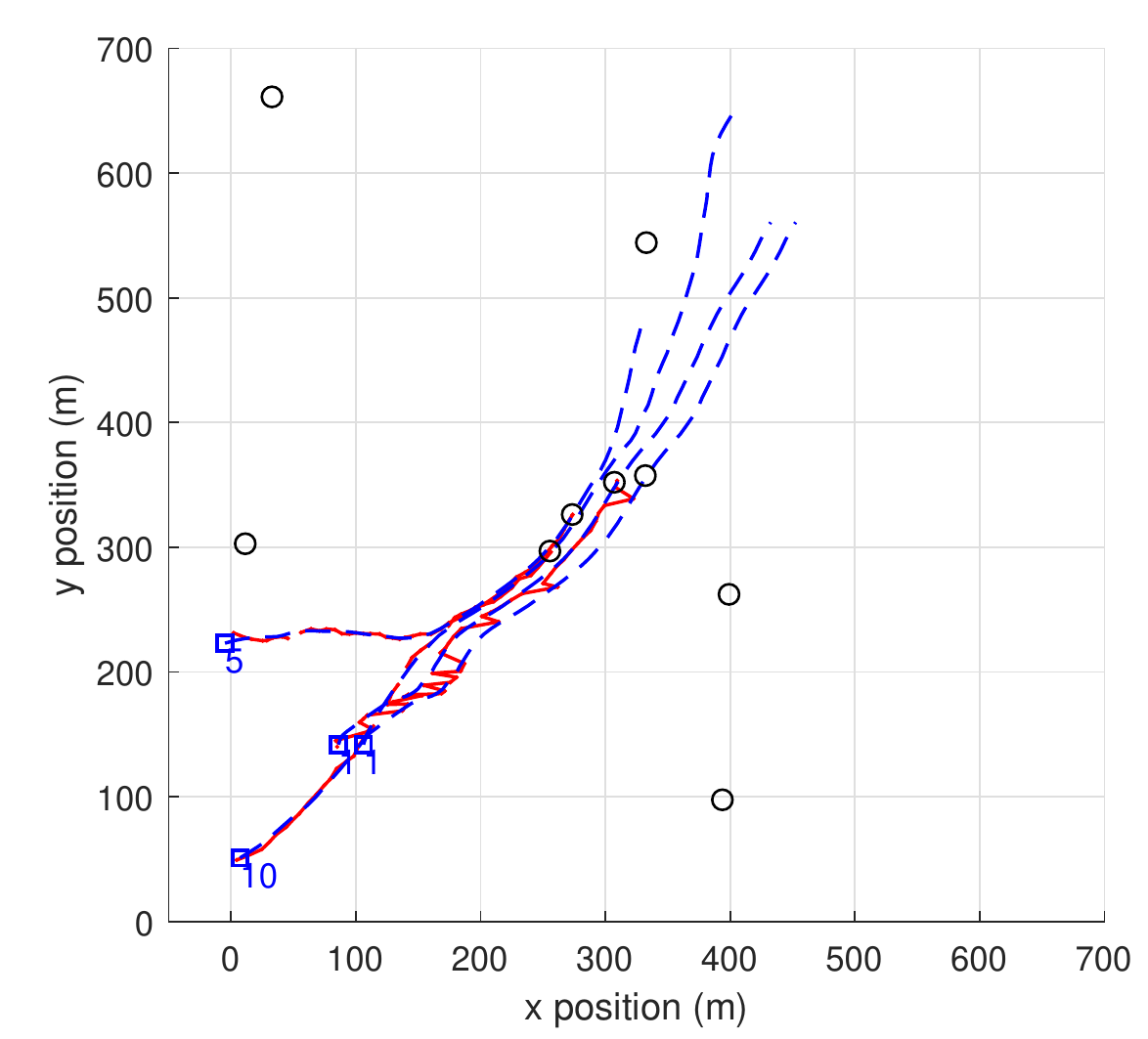}\includegraphics[scale=0.37]{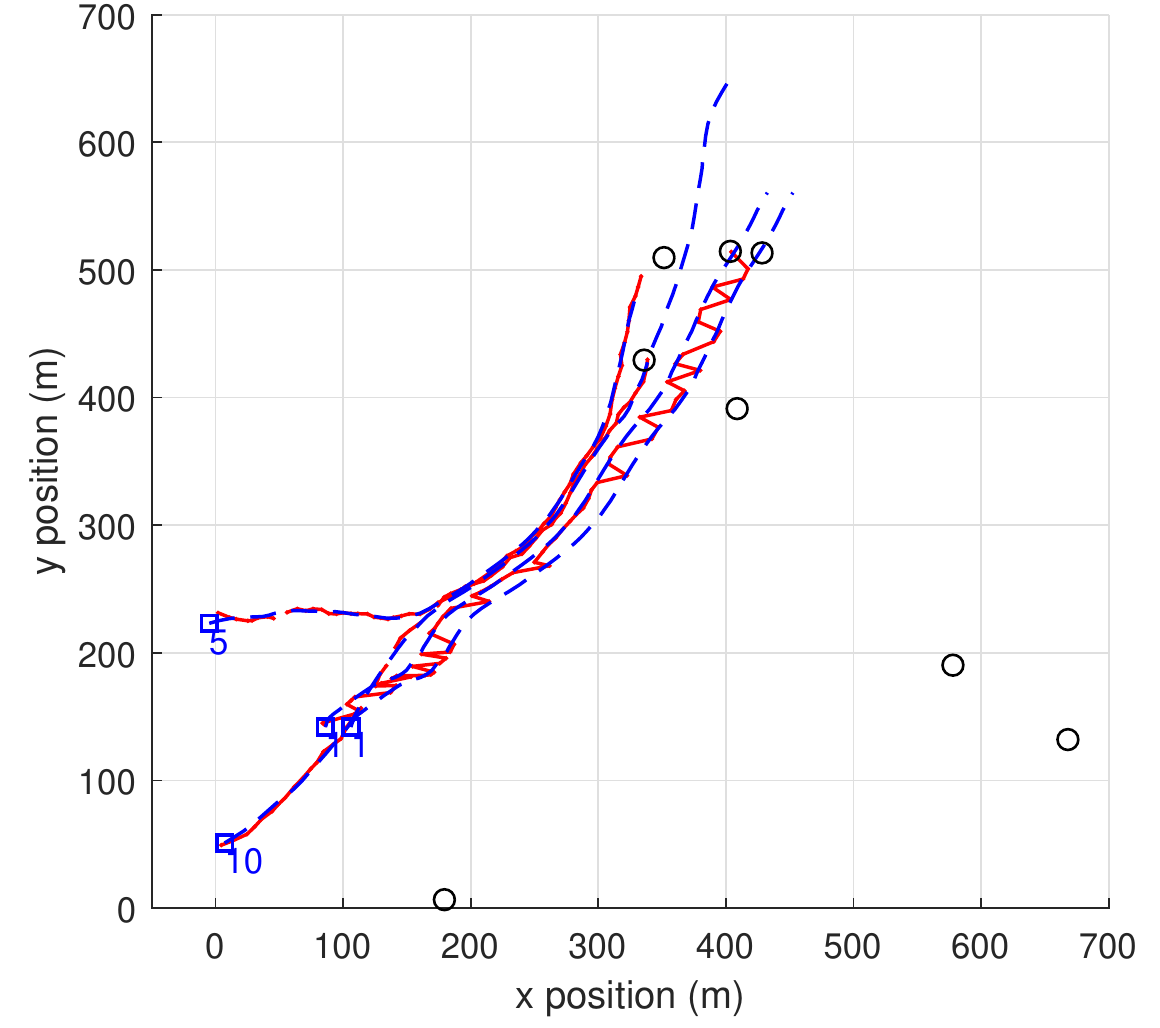}
\par\end{centering}
\caption{\label{fig:Exemplar-outputs}Exemplar outputs of the TPHD filter (top)
and tagged PHD filter (bottom) at time steps 50 (left) and 70 (right),
shown in a subregion of the surveillance area. The dashed blue lines
represent the true trajectories. The blue squares and the numbers
next to them denote the starting positions and starting times of different
trajectories. There are four trajectories, two start at time step
1, close to each other, and end at time step 79. The other two start
at time steps 5 and 10 and end at time steps 69 and 94, respectively.
Black circles represent the current measurements. The TPHD filter
is able to estimate the alive trajectories at each time step. The
tagged PHD filter does not work well, as the estimator selects the
highest peaks with distinct tags and there are two targets born from
the same PHD component at the same time, which leads to track switching,
false and missed targets. At time step 50, the tagged PHD filter estimates
four trajectories, with one of length one in the birth location $\left[85,140\right]^{T}$.
The tagged PHD filter at time step 70 estimates three trajectories,
though one of them is no longer alive.}
\end{figure}

We have implemented the $L$-scan GMTPHD and GMTCPHD filters with
$L\in\left\{ 1,2,5\right\} $ in a scenario with $N_{s}=100$ time
steps. We use a pruning threshold $\Gamma_{p}=10^{-4}$, absorption
threshold $\Gamma_{a}=4$ and limit the number of components to 30.
An exemplar output of the $1$-scan TPHD filter, the tagged PHD filter
and the considered ground truth are shown in Figure \ref{fig:Exemplar-outputs}.
At each time step, TPHD and TCPHD filters provide an estimate of the
set of present trajectories at the current time. The tagged PHD filter
shows considerably worse performance as two targets are born at the
same time step from the same PHD component. The start and end times
of an estimated trajectory do not depend on the choice of $L$ so
the output for any other $L$ has the same start time and duration,
but with a different error. 

In the following, we evaluate the performance of the filters by Monte
Carlo simulation with $N_{mc}=500$ runs. At each time step $k$,
we measure the error between the set $\mathbf{X}_{a}^{k}$ of alive
trajectories and its estimate $\mathbf{\hat{X}}_{a}^{k}$. In order
to do so, we use the metric for sets of trajectories based on linear
programming in \cite{Rahmathullah16_prov2} with parameters $p=2$,
$c=10$ and $\gamma=1$, which we denote here as $d\left(\cdot,\cdot\right)$.
In this case, $d^{2}\left(\cdot,\cdot\right)$ can be decomposed into
the square costs: $c_{m}^{2}\left(\cdot,\cdot\right)$ for missed
targets, $c_{f}^{2}\left(\cdot,\cdot\right)$ for false targets, $c_{l}^{2}\left(\cdot,\cdot\right)$
for the localisation error of properly detected targets, and $c_{t}^{2}\left(\cdot,\cdot\right)$
for track switches, as in performance evaluation in traditional multitarget
tracking \cite[Sec. 13.6]{Blackman_book99}. That is, we have
\begin{align}
d^{2}\left(\mathbf{X}_{a}^{k},\mathbf{\hat{X}}_{a}^{k}\right) & =c_{l}^{2}\left(\mathbf{X}_{a}^{k},\mathbf{\hat{X}}_{a}^{k}\right)+c_{m}^{2}\left(\mathbf{X}_{a}^{k},\mathbf{\hat{X}}_{a}^{k}\right)\nonumber \\
 & \quad+c_{f}^{2}\left(\mathbf{X}_{a}^{k},\mathbf{\hat{X}}_{a}^{k}\right)+c_{t}^{2}\left(\mathbf{X}_{a}^{k},\mathbf{\hat{X}}_{a}^{k}\right).\label{eq:metric_decomposition}
\end{align}
This decomposition is useful to analyse the performances of the filters,
as done in this section.  In our results, we only use the position
elements to compute the error and normalise the error by the considered
time window such that the squared error at time $k$ is $d^{2}\left(\mathbf{X}_{a}^{k},\mathbf{\hat{X}}_{a}^{k}\right)/k$.
The root mean square (RMS) error at a given time step is calculated
as
\begin{align}
d\left(k\right) & =\sqrt{\frac{1}{N_{mc}k}\sum_{i=1}^{N_{mc}}d^{2}\left(\mathbf{X}_{a}^{k},\mathbf{\hat{X}}_{a,i}^{k}\right)},\label{eq:error_time_k}
\end{align}
where $\mathbf{\hat{X}}_{a,i}^{k}$ is the estimate of the alive trajectories
at time $k$ in the $i$th Monte Carlo run. In this scenario, the
TPHD and TCPHD filters estimates do not have track switches and the
resulting errors using $d\left(\mathbf{X}_{a}^{k},\mathbf{\hat{X}}_{a,i}^{k}\right)$
are the same as the errors computed by the sum of the generalised
optimal sub-pattern assignment (GOSPA) metric ($\alpha=2$) \cite{Rahmathullah17}
between the true targets and their estimates across all time steps.
The tagged PHD and CPHD filters show track switches and therefore,
the trajectory metric and GOSPA provide different errors.

\begin{figure}
\begin{centering}
\includegraphics[scale=0.6]{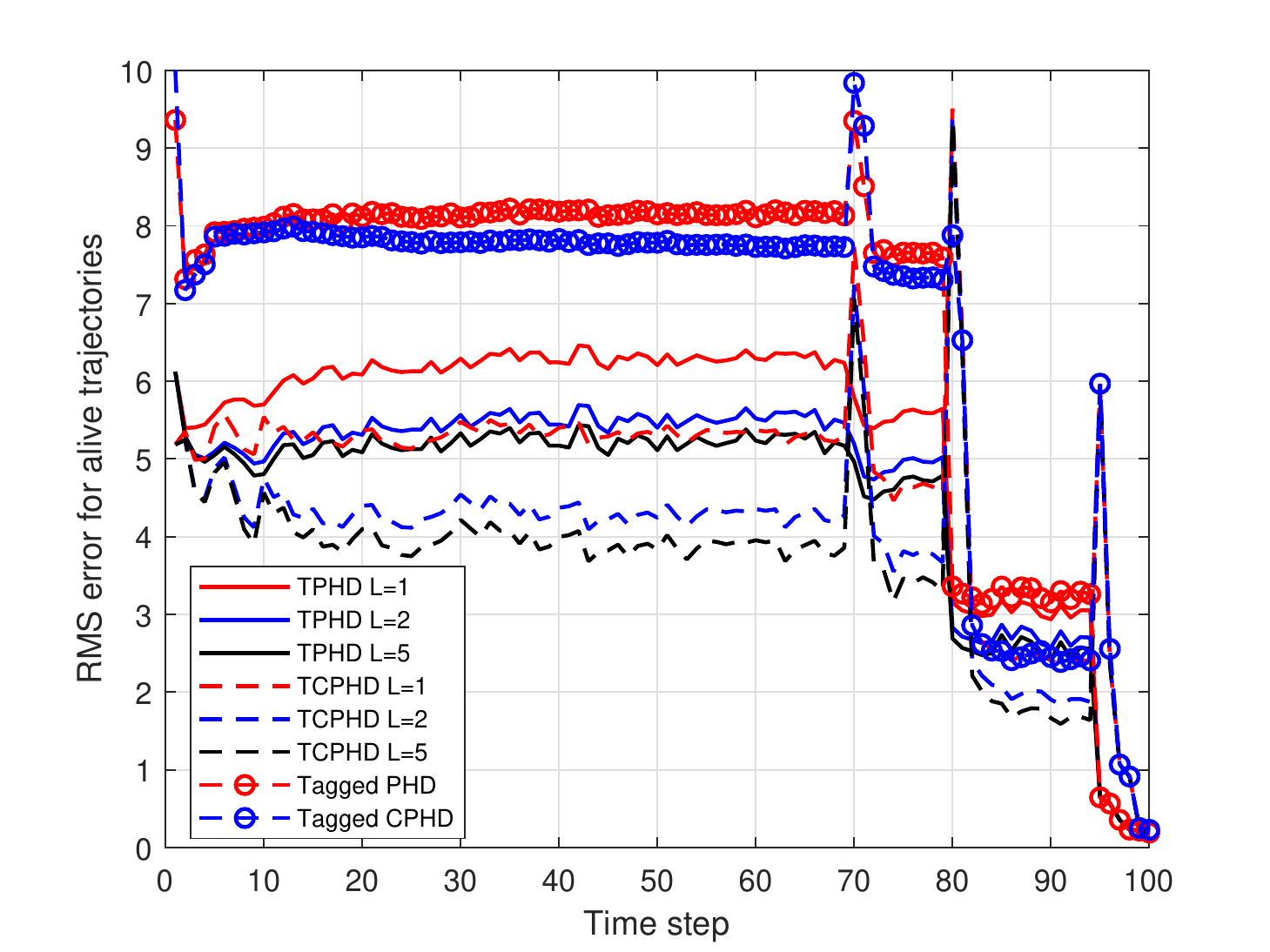}
\par\end{centering}
\caption{\label{fig:Error_trajectories}RMS trajectory metric error (\ref{eq:error_time_k})
of the alive trajectories for the TPHD/TCPHD filter and tagged PHD/CPHD
filters. Filters based on sets of trajectories have a much higher
performance than tagged filters. Increasing $L$ lowers the error.
}
\end{figure}

\begin{figure}
\begin{centering}
\includegraphics[scale=0.6]{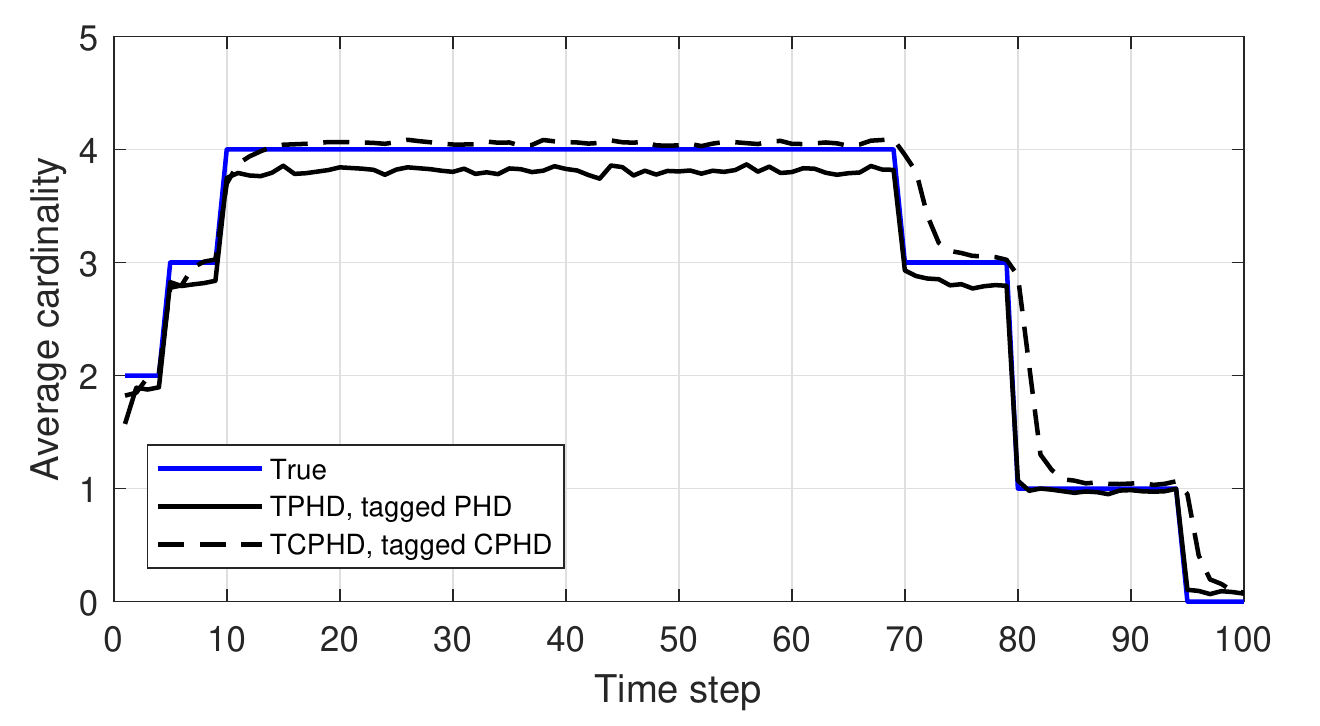}
\par\end{centering}
\caption{\label{fig:Estimated-cardinality}Estimated cardinality against time
for the filters. CPHD-based filters estimate the cardinality more
accurately except just after a target disappears.}
\end{figure}

The RMS trajectory errors for the algorithms are plotted in Figure
\ref{fig:Error_trajectories}. As expected, for the TPHD and TCPHD,
increasing $L$ improves estimation performance and lowers the error.
The TCPHD filter shows lower errors than the TPHD filter, except when
targets disappear. The TCPHD filter also estimates the cardinality
more accurately than the TPHD filter, except just after a target disappears,
as can be seen in Figure \ref{fig:Estimated-cardinality}. The tagged
PHD and CPHD filters provide a considerably higher error, as track
estimation is done in a manner that does not work well if two targets
are born from the same PHD component at the same time.

\begin{figure}
\begin{centering}
\includegraphics[scale=0.6]{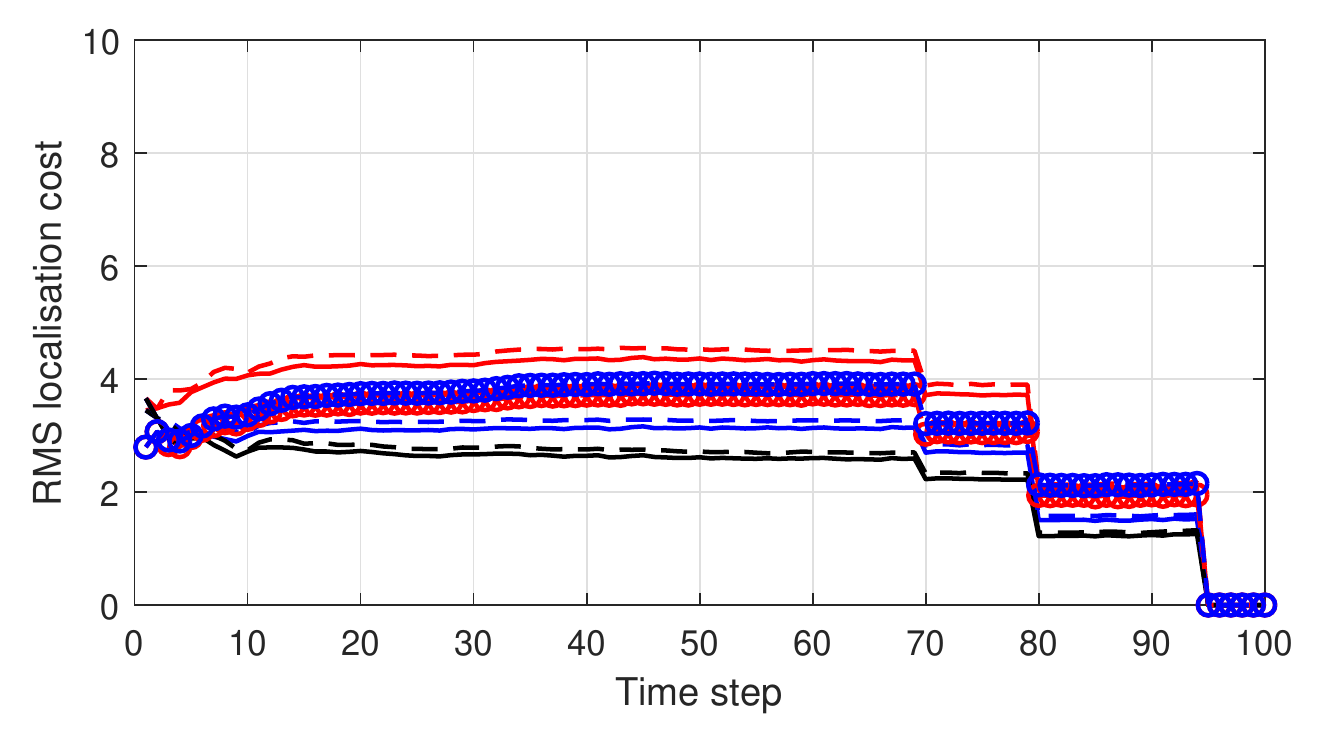}
\par\end{centering}
\begin{centering}
\includegraphics[scale=0.6]{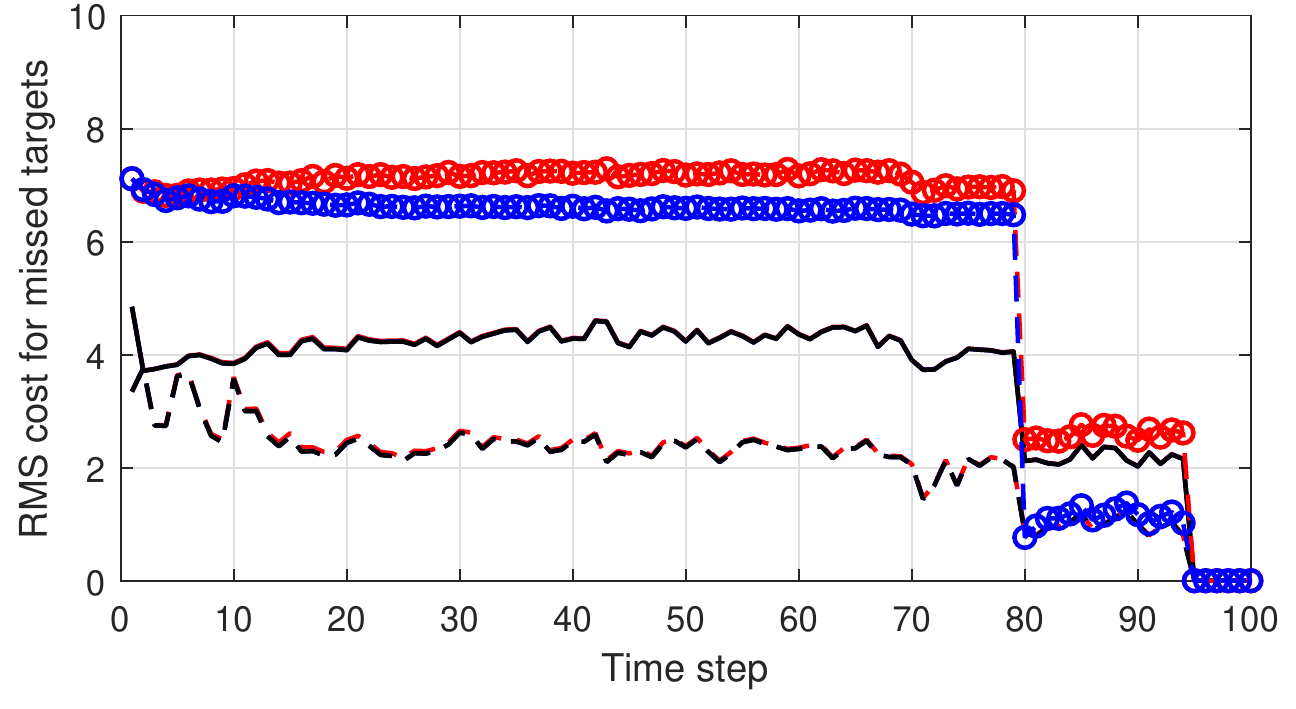}
\par\end{centering}
\begin{centering}
\includegraphics[scale=0.6]{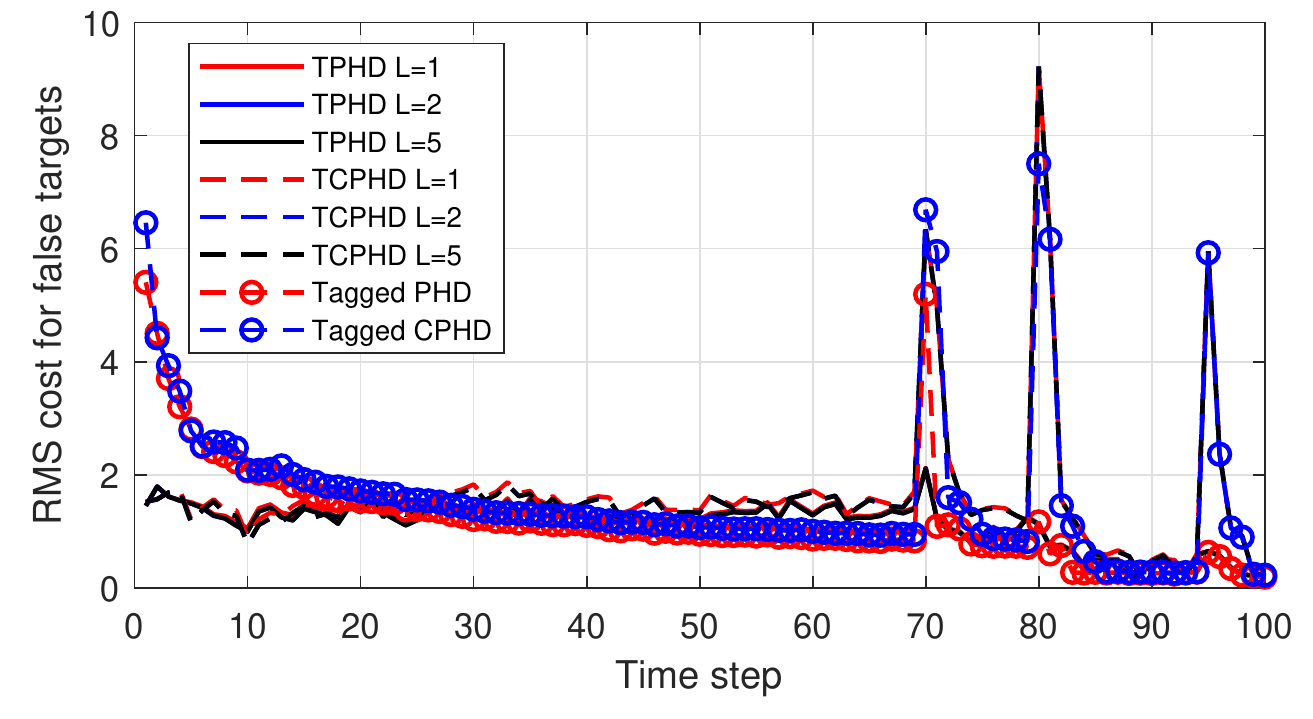}
\par\end{centering}
\begin{centering}
\includegraphics[scale=0.6]{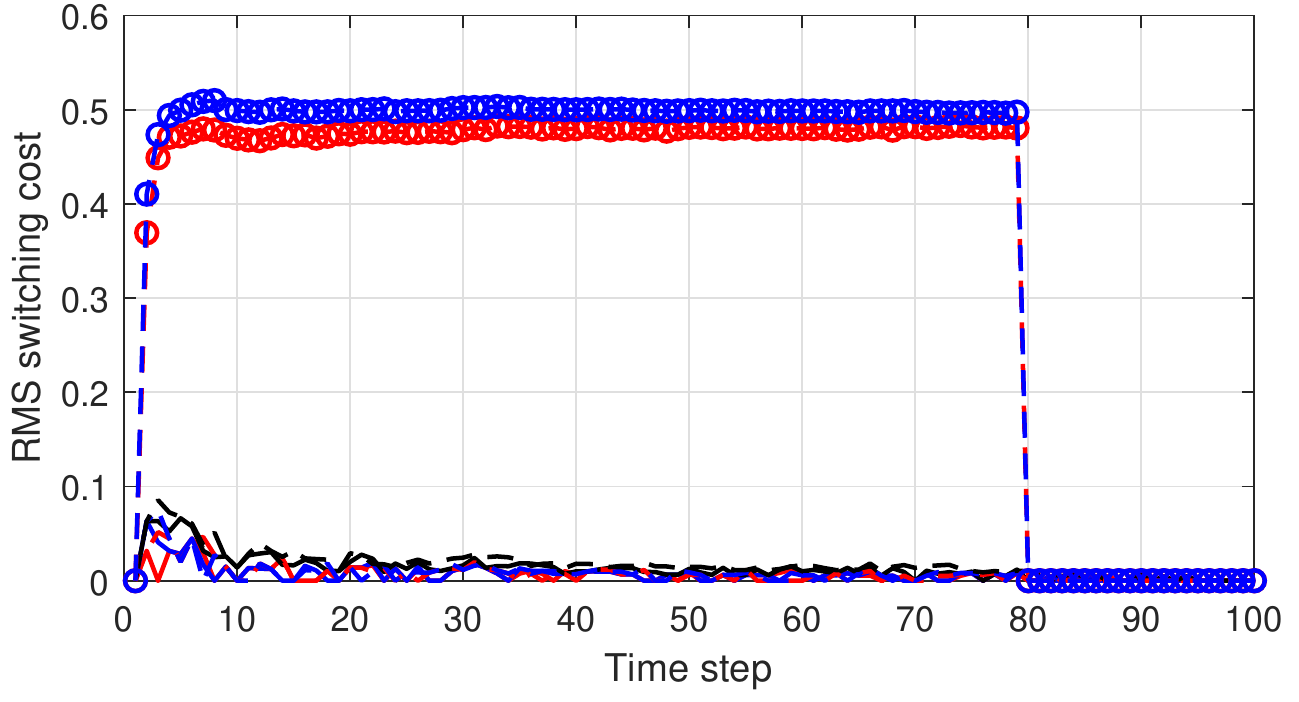}
\par\end{centering}
\caption{\label{fig:Metric_decomposition_cost}RMS costs for localisation errors
for properly detected targets, missed targets, false targets and track
switches for the alive trajectories at every time step. The scale
of the $y$ axis changes in the figure for the switching cost. The
RMS costs for missed targets and false targets do not change with
the considered value of $L$. Increasing $L$, decreases the localisation
cost for trajectory-based algorithms. TCPHD misses fewer targets than
TPHD. Switching cost is negligible for TPHD/TCPHD filters. The tagged
PHD and CPHD filters have a high cost for missed targets and show
track switching.}
\end{figure}

In order to analyse the results more thoroughly, we make use of the
metric decomposition indicated in (\ref{eq:metric_decomposition}).
We compute the RMS costs in (\ref{eq:metric_decomposition}) at each
time step normalised by $k$, as in (\ref{eq:error_time_k}). The
results are shown in Figure \ref{fig:Metric_decomposition_cost}.
Tagged PHD and CPHD filters have a higher cost for missed targets
and track switches than TPHD and TCPHD filters, as there are two targets
born from the same PHD component and tagging does not work well in
this case. We can see that increasing $L$ does not change the costs
for missed and false targets and track switches for the trajectory
algorithms, it mainly improves the localisation costs. We can see
that the main advantage of using the TCPHD filter over the TPHD filter
is the reduction of the number of missed targets. Both filters behave
quite similarly in terms of localisation costs, missed targets and
track switches. We can also see that the TCPHD filter has a higher
cost for false targets than the PHD filter at time steps 70, 80 and
95, just after a target disappears. 

The running times of our Matlab implementations of the filters in
a computer with a 3.5 GHz Intel Xeon E5 processor are shown in Table
\ref{tab:Running_times}. The running times of the TCPHD filter are
roughly 1 second longer than the TPHD for all values of $L$. For
visualisation clarity, we have only shown performance results for
$L\in\left\{ 1,2,5\right\} $ in Figures \ref{fig:Error_trajectories}
and \ref{fig:Metric_decomposition_cost}. For any $L\in\left\{ 10,20,30\right\} $,
performance is similar and slightly better than for $L=5$. In particular,
the RMS error (\ref{eq:error_time_k}) considering all time steps
\begin{align}
d_{T}= & \sqrt{1/N_{s}\cdot\sum_{k=1}^{N_{s}}d^{2}\left(k\right)}\label{eq:RMS_error_all_time}
\end{align}
using the trajectory metric is decreased from 4.68 (TPHD) and 3.90
(TCPHD) with $L=5$ to 4.66 (TPHD) and 3.87 (TCPHD) with $L\in\left\{ 10,20,30\right\} $.
Tagged filters have a higher computational burden than TPHD and TCPHD
with low $L$, due to the estimation process that links the current
target state estimates with the previously estimated trajectories
using the tags. 

\begin{table}
\caption{\label{tab:Running_times}Running times of the algorithms in seconds}
\centering{}%
\begin{tabular}{l|llllll}
\hline 
$L$ &
1  &
2 &
5 &
10 &
20 &
30\tabularnewline
\hline 
TPHD &
1.1 &
1.1 &
1.2 &
1.7 &
3.4 &
6.0\tabularnewline
TCPHD &
2.0 &
2.0 &
2.1 &
2.6 &
4.3 &
6.9\tabularnewline
\hline 
Tagged PHD &
\multicolumn{6}{c}{2.2}\tabularnewline
Tagged CPHD &
\multicolumn{6}{c}{3.2}\tabularnewline
\hline 
\end{tabular}
\end{table}

\begin{table*}
\caption{\label{tab:Error-TPHD-TCPHD}Error in alive trajectories averaged
over all time steps for the trajectory metric (TM) and OSPA}
\begin{centering}
\par\end{centering}
\begin{centering}
\par\end{centering}
\centering{}%
\begin{tabular}{c|cccccc|cccccc|cc|cc}
\hline 
 &
\multicolumn{6}{c|}{TPHD} &
\multicolumn{6}{c|}{TCPHD} &
\multicolumn{2}{c|}{ Tagged PHD } &
\multicolumn{2}{c}{Tagged CPHD}\tabularnewline
\hline 
 &
\multicolumn{3}{c}{TM/GOSPA} &
\multicolumn{3}{c|}{OSPA} &
\multicolumn{3}{c}{TM/GOSPA} &
\multicolumn{3}{c|}{OSPA} &
TM &
OSPA &
TM &
OSPA\tabularnewline
\hline 
$L$ &
1 &
2 &
5 &
1 &
2 &
5 &
1 &
2 &
5 &
1 &
2 &
5 &
- &
- &
- &
-\tabularnewline
No change &
5.54 &
4.89 &
4.68 &
3.94 &
3.64 &
3.55 &
4.98 &
4.17 &
\uline{3.90} &
3.39 &
3.01 &
\uline{2.89} &
7.31 &
5.55 &
7.12 &
5.34\tabularnewline
$\sigma^{2}=16$ &
9.31 &
8.50 &
8.04 &
5.83 &
5.40 &
5.17 &
8.27 &
8.39 &
\uline{7.88} &
5.61 &
5.14 &
\uline{4.87} &
8.95 &
6.37 &
8.94 &
6.29\tabularnewline
$\sigma^{2}=1$ &
4.48 &
4.18 &
4.13 &
3.43 &
3.30 &
3.28 &
3.71 &
3.30 &
\uline{3.23} &
2.77 &
2.59 &
\uline{2.56} &
6.81 &
5.33 &
6.55 &
5.09\tabularnewline
$\lambda_{c}=70$ &
5.61 &
4.98 &
4.78 &
4.01 &
3.72 &
3.63 &
5.09 &
4.33 &
\uline{4.07} &
3.50 &
3.15 &
\uline{3.03} &
7.27 &
5.53 &
7.05 &
5.30\tabularnewline
$\lambda_{c}=30$ &
5.49 &
4.84 &
4.63 &
3.90 &
3.60 &
3.51 &
4.89 &
4.08 &
\uline{3.81} &
3.31 &
2.93 &
\uline{2.81} &
7.28 &
5.52 &
7.06 &
5.29\tabularnewline
$p_{D}=0.99$ &
4.15 &
3.34 &
3.09 &
2.63 &
2.21 &
2.09 &
4.07 &
3.23 &
\uline{2.97} &
2.57 &
2.14 &
\uline{2.01} &
7.17 &
5.36 &
7.20 &
5.35\tabularnewline
$p_{D}=0.85$ &
5.88 &
5.19 &
4.93 &
4.15 &
3.84 &
3.73 &
5.48 &
4.68 &
\uline{4.39} &
3.80 &
3.43 &
\uline{3.30} &
7.28 &
5.54 &
7.15 &
5.38\tabularnewline
$p_{D}=0.75$ &
6.77 &
6.07 &
5.76 &
4.84 &
4.53 &
4.41 &
6.40 &
5.59 &
\uline{5.23} &
4.47 &
4.11 &
\uline{3.95} &
7.59 &
5.81 &
7.35 &
5.55\tabularnewline
$p_{S}=0.95$  &
5.57 &
4.93 &
4.72 &
3.97 &
3.68 &
3.59 &
5.08 &
4.31 &
\uline{4.06} &
3.49 &
3.12 &
\uline{3.01} &
7.35 &
5.60 &
7.16 &
5.38\tabularnewline
$w_{\beta,j}^{k}=0.05$ &
5.56 &
4.91 &
4.70 &
3.97 &
3.68 &
3.59 &
5.00 &
4.21 &
\uline{3.94} &
3.45 &
3.08 &
\uline{2.96} &
7.33 &
5.58 &
7.15 &
5.40\tabularnewline
\hline 
\end{tabular}
\end{table*}

We also show the RMS error (\ref{eq:RMS_error_all_time}) for other
simulation parameters in Table \ref{tab:Error-TPHD-TCPHD}. In this
table, we also include results where $d\left(\cdot,\cdot\right)$
is the sum of the OSPA error \cite{Schuhmacher08_b,Schuhmacher08},
with the same $p$ and $c$, between target states and their estimates
at each time step. The trajectory metric and GOSPA return the same
errors for TPHD/TCPHD filters, for all simulation parameters considered,
which implies that the cost of track switches is negligible. Tagged
filters show track switches so the trajectory metric and GOSPA do
not coincide. GOSPA errors are not shown due to space constraints,
but they are generally 0.02 lower than the trajectory metric error
for the tagged filters. According to all the metrics and all scenarios,
the TCPHD filter is the best performing filter, followed by the TPHD
filter. For both filters, all metrics and scenarios, the error decreases
as $L$ increases. The errors of the tagged filters are, in general,
considerably higher than for the set of trajectories filters. As expected,
performance improves for all filters if the probability of detection
increases, clutter intensity decreases and the measurement noise variance
decreases. We have also shown results with a different probability
of survival and birth intensity to show that the results are consistent
with different simulation parameters.

\section{Conclusions\label{sec:Conclusions}}

In this paper we have developed the trajectory PHD and CPHD filters
based on KLD minimisations and sets of trajectories. The TPHD and
TCPHD filters propagate a Poisson multitrajectory density and an IID
cluster multitrajectory density through the filtering recursion to
make inference on the set of alive trajectories. The theory presented
in this paper endows the PHD/CPHD filters with the capability of estimating
trajectories from first principles, which can span their already widespread
use to more applications where tracks are required.

We have also proposed a Gaussian mixture implementation of the filters.
In particular, the parameter $L$ of the $L$-scan versions governs
the accuracy of the estimation of past states of the trajectories.
We have analysed the proposed filters in terms of localisation errors,
missed targets, false targets and track switches, and also based on
OSPA. Increasing $L$ mainly improves the localisation error of past
states of the trajectory for both filters. In general, the TCPHD filter
outperforms the TPHD filter, and both provide better trajectory estimates
than the ones provided by tagging the PHD and CPHD filters.

\bibliographystyle{IEEEtran}
\bibliography{20E__Trabajo_Angel_Mis_articulos_Finished_Trajectory_PHD-CPHD_filter_Accepted_Referencias}

\cleardoublepage{}
\begin{center}
{\LARGE{}Supplementary material of ``Trajectory PHD and CPHD filters'' }
\par\end{center}{\LARGE \par}

\appendices{}

\section{\label{sec:Appendix_sampling}}

In this appendix, we indicate how to draw samples from an IID cluster
trajectory RFS, which includes the Poisson trajectory RFS as a particular
case. Given a single trajectory density $\breve{\nu}\left(\cdot\right)$,
the probability that the trajectory starts at time $t$ and has duration
$i$ is 
\begin{equation}
P_{\breve{\nu}}\left(t,i\right)=\int\breve{\nu}\left(t,x^{1:i}\right)dx^{1:i}.\label{eq:prob_start_time_duration}
\end{equation}
Given the start time $t$ and duration $i$, the density of the target
states of the trajectory is
\begin{align}
\breve{\nu}\left(x^{1:i}|t,i\right) & =\frac{\breve{\nu}\left(t,x^{1:i}\right)}{P_{\breve{\nu}}\left(t,i\right)}.\label{eq:conditional_pdf_IID}
\end{align}
Therefore, we can draw samples from an IID cluster trajectory RFS
using (\ref{eq:prob_start_time_duration}) and (\ref{eq:conditional_pdf_IID}),
as indicated in Algorithm \ref{alg:Sampling_IID_cluster}.

\begin{algorithm}
\caption{\label{alg:Sampling_IID_cluster}Sampling from an IID cluster multitrajectory
density}

{\fontsize{9}{9}\selectfont

\textbf{Input:} IID cluster multitrajectory density $\nu\left(\cdot\right)$. 

\textbf{Output: }Sample $X\thicksim\nu\left(\cdot\right)$.

\begin{algorithmic}     

\State - Set $X=\emptyset$ and sample $n\thicksim\rho_{\nu}\left(\cdot\right)$. 

\For{ $j=1$ to $n$ } 

\State - Sample $\left(t,i\right)\thicksim P_{\breve{\nu}}\left(\cdot\right)$
and $x^{1:i}\thicksim\breve{\nu}\left(\cdot|t,i\right)$, see (\ref{eq:prob_start_time_duration})
and (\ref{eq:conditional_pdf_IID}).

\State - Set $X\leftarrow X\cup\left\{ \left(t,x^{1:i}\right)\right\} $. 

\EndFor

\end{algorithmic}

}
\end{algorithm}

\section{\label{sec:Appendix_KLD}}

In this appendix we prove Theorem \ref{thm:KLD_minimisation_iidc}.
A multitrajectory density $\pi\left(\cdot\right)$ can be written
as \cite{Mahler03}
\begin{equation}
\pi\left(\left\{ X_{1},...,X_{n}\right\} \right)=\rho_{\pi}\left(n\right)n!\pi_{n}\left(X_{1},...,X_{n}\right)\label{eq:appendix_general_density}
\end{equation}
where $\pi_{n}\left(\cdot\right)$ is a permutation invariant ordered
trajectory density such that
\begin{align*}
\int\pi_{n}\left(X_{1},...,X_{n}\right)dX_{1:n} & =1.
\end{align*}
The marginal density of this density is written as
\begin{align*}
\tilde{\pi}_{n}\left(X\right) & =\int\pi_{n}\left(X,X_{2}...,X_{n}\right)dX_{2:n}\\
 & =\frac{1}{\rho_{\pi}\left(n\right)n!}\int\pi\left(\left\{ X,X_{2},...,X_{n}\right\} \right)dX_{2:n}.
\end{align*}
Substituting (\ref{eq:iidc_prior}) and (\ref{eq:appendix_general_density})
into (\ref{eq:KLD_definition}), we get
\begin{align}
\mathrm{D}\left(\pi\left\Vert \nu\right.\right) & =\sum_{n=0}^{\infty}\rho_{\pi}\left(n\right)\log\frac{\rho_{\pi}\left(n\right)}{\rho_{\nu}\left(n\right)}\nonumber \\
 & \quad+\sum_{n=0}^{\infty}\rho_{\pi}\left(n\right)\int\pi_{n}\left(X_{1},...,X_{n}\right)\nonumber \\
 & \quad\times\log\frac{\pi_{n}\left(X_{1},...,X_{n}\right)}{\prod_{j=1}^{n}\breve{\nu}\left(X_{j}\right)}dX_{1:n}.\label{eq:append_KLD_min}
\end{align}
The objective is to find $\rho_{\nu}\left(\cdot\right)$ and $\breve{\nu}\left(\cdot\right)$
that minimise $\mathrm{D}\left(\pi\left\Vert \nu\right.\right)$.
From KLD minimisation over discrete variables,  $\rho_{\nu}\left(\cdot\right)=\rho_{\pi}\left(\cdot\right)$
minimises the KLD. Minimizing $\mathrm{D}\left(\pi\left\Vert \nu\right.\right)$
w.r.t. $\breve{\nu}\left(\cdot\right)$ is equivalent to minimising
the functional
\begin{align*}
L\left[\breve{\nu}\right] & =-\int\sum_{n=0}^{\infty}\rho_{\pi}\left(n\right)n\tilde{\pi}_{n}\left(X\right)\log\breve{\nu}\left(X\right)dX.
\end{align*}
which is minimised by \cite{Angel18_c} 
\begin{align*}
\breve{\nu}\left(X\right) & =\frac{D_{\pi}\left(X\right)}{\sum_{n=0}^{\infty}\rho_{\pi}\left(n\right)n},
\end{align*}
or equivalently, $D_{\nu}\left(\cdot\right)=D_{\pi}\left(\cdot\right)$.

\section{\label{sec:IIDcluster_current_time}}

In this appendix, we prove (\ref{eq:cardinality_alive_IID_example}).
The IID cluster multitrajectory density $\nu^{k}\left(\cdot\right)$
that minimises the KLD has the same cardinality distribution and PHD
as $\pi^{k}\left(\cdot\right)$. The cardinality distribution of $\pi^{k}\left(\cdot\right)$
is such that $\rho_{\pi^{k}}\left(m\right)=1$, so we compute its
PHD. The posterior can be written as \cite[Eq. (12.91)]{Mahler_book07}
\begin{align}
\pi^{k}\left(\left\{ X_{1},...,X_{m}\right\} \right) & =\sum_{\sigma\in\Xi_{m}}\left[\prod_{i=1}^{m_{a}}\breve{a}_{i}\left(X_{\sigma_{i}}\right)\right]\left[\prod_{i=1}^{m_{d}}\breve{d}_{i}\left(X_{\sigma_{i+m_{a}}}\right)\right]\label{eq:posterior_example}
\end{align}
where $\Xi_{m}$ is the set that contains all permutations of $\left(1,...,m\right)$
and $\sigma_{i}$ is the $i$th component of $\sigma$.

Substituting (\ref{eq:posterior_example}) into (\ref{eq:PHD}), the
PHD of the posterior is
\begin{align*}
D_{\pi^{k}}(X) & =\sum_{n=0}^{\infty}\frac{1}{n!}\int\pi^{k}\left(\left\{ X,X_{1},...,X_{n}\right\} \right)dX_{1:n}\\
 & =\sum_{i=1}^{m_{a}}\breve{a}_{i}\left(X\right)+\sum_{i=1}^{m_{d}}\breve{d}_{i}\left(X\right),
\end{align*}
which is equal to $D_{\nu^{k}}(X)$. Then, using (\ref{eq:PHD_IID_cluster}),
we have that the single trajectory density $\breve{\nu}^{k}\left(\cdot\right)$
is 
\begin{align*}
\breve{\nu}^{k}\left(X\right) & =\frac{D_{\nu^{k}}(X)}{m}\\
 & =\frac{\sum_{i=1}^{m_{a}}\breve{a}_{i}\left(X\right)+\sum_{i=1}^{m_{d}}\breve{d}_{i}\left(X\right)}{m}.
\end{align*}
A single trajectory with this density exists at the current time $k$
with probability $m_{a}/m$. As we have $m$ IID trajectories with
this distribution, the number of alive trajectories follows the binomial
distribution in (\ref{eq:cardinality_alive_IID_example}).

\section{\label{sec:Appendix_TCPHD_update}}

In this appendix, we prove Theorem \ref{thm:TCPHD_update} (TCPHD
update). The proof is quite similar to the CPHD update \cite{Angel15_d}. 

\subsection{Density of the measurement}

As the current set of targets is an IID cluster with cardinality $\rho_{\omega^{k}}\left(\cdot\right)$
and PHD $D_{\omega_{T}^{k}}\left(\cdot\right)$, the density of the
measurements, under Assumptions U1, U4 and U5, is the same as in the
CPHD filter \cite[Eq. (32)]{Angel15_d}. 
\begin{align}
\ell^{k}\left(\mathbf{z}^{k}\right) & =\left\langle \rho_{\omega^{k}},\Upsilon^{0}\left[D_{\omega_{\tau}^{k}},\mathbf{z}^{k}\right]\right\rangle \prod_{z\in\mathbf{z}^{k}}\breve{c}\left(z\right).\label{eq:density_measurements_CPHD}
\end{align}

\subsection{Cardinality of the posterior}

As the set of trajectories present at the current time has the same
cardinality as the set of targets at the current time, the updated
cardinality distribution is the same in both cases so the TCPHD cardinality
update is the same as the CPHD filter cardinality update, which is
given in (\ref{eq:cardinality_update_CPHD}).

\subsection{PHD of the posterior}

The PHD of the posterior is
\begin{align*}
 & D_{\pi^{k}}(X)\\
 & \quad=\int\pi^{k}\left(\left\{ X\right\} \cup\mathbf{X}\right)\delta\mathbf{X}\\
 & \quad=\frac{1}{\ell^{k}\left(\mathbf{z}^{k}\right)}\int\ell^{k}\left(\mathbf{z}^{k}\left|\tau^{k}\left(\left\{ X\right\} \cup\mathbf{X}\right)\right.\right)\omega^{k}\left(\left\{ X\right\} \cup\mathbf{X}\right)\delta\mathbf{X}\\
 & \quad=\frac{1}{\ell^{k}\left(\mathbf{z}^{k}\right)}\sum_{n=0}^{\infty}\left(n+1\right)\int\ell^{k}\left(\mathbf{z}^{k}|\tau^{k}\left(\left\{ X,X_{1},...,X_{n}\right\} \right)\right)\\
 & \quad\times\rho_{\omega^{k}}\left(n+1\right)\breve{\omega}^{k}\left(X\right)\left[\prod_{j=1}^{n}\breve{\omega}^{k}\left(X_{j}\right)\right]dX_{1:n}.
\end{align*}

We apply the decomposition \cite[Eq. (14)]{Angel15_d} 
\begin{align*}
 & \ell^{k}\left(\mathbf{z}^{k}\left|\tau^{k}\left(\left\{ X\right\} \right)\cup\tau^{k}\left(\mathbf{X}\right)\right.\right)\\
 & \quad=\left(1-p_{D}\left(\tau^{k}\left(\left\{ X\right\} \right)\right)\right)\ell^{k}\left(\mathbf{z}^{k}\left|\tau^{k}\left(\mathbf{X}\right)\right.\right)\\
 & \quad+p_{D}\left(\tau^{k}\left(\left\{ X\right\} \right)\right)\sum_{z\in\mathbf{z}^{k}}l\left(z|\tau^{k}\left(\left\{ X\right\} \right)\right)\ell^{k}\left(\mathbf{z}^{k}\setminus\left\{ z\right\} \left|\tau^{k}\left(\mathbf{X}\right)\right.\right)
\end{align*}
and obtain $D_{\pi^{k}}(t,x^{1:i})$

\begin{align}
 & =\frac{\left(1-p_{D}\left(x^{i}\right)\right)\breve{\omega}^{k}\left(t,x^{1:i}\right)}{\ell^{k}\left(\mathbf{z}^{k}\right)}\sum_{n=0}^{\infty}\left(n+1\right)\rho_{\omega^{k}}\left(n+1\right)\nonumber \\
 & \quad\times\int\ell^{k}\left(\mathbf{z}^{k}|\tau^{k}\left(\left\{ X_{1},...,X_{n}\right\} \right)\right)\left[\prod_{j=1}^{n}\breve{\omega}^{k}\left(X_{j}\right)\right]dX_{1:n}\nonumber \\
 & \quad+\frac{p_{D}\left(x^{i}\right)\breve{\omega}^{k}\left(t,x^{1:i}\right)}{\ell^{k}\left(\mathbf{z}^{k}\right)}\sum_{z\in\mathbf{z}^{k}}l\left(z|x^{i}\right)\sum_{n=0}^{\infty}\left(n+1\right)\rho_{\omega^{k}}\left(n+1\right)\nonumber \\
 & \quad\times\int\ell^{k}\left(\mathbf{z}^{k}\setminus\left\{ z\right\} \left|\tau^{k}\left(\left\{ X_{1},...,X_{n}\right\} \right)\right.\right)\left[\prod_{j=1}^{n}\breve{\omega}^{k}\left(X_{j}\right)\right]dX_{1:n}\nonumber \\
 & =\frac{\left(1-p_{D}\left(x^{i}\right)\right)\breve{\omega}^{k}\left(t,x^{1:i}\right)}{\ell^{k}\left(\mathbf{z}^{k}\right)}\sum_{n=0}^{\infty}\left(n+1\right)\rho_{\omega^{k}}\left(n+1\right)\nonumber \\
 & \quad\times\int\ell^{k}\left(\mathbf{z}^{k}|\left\{ x_{1},...,x_{n}\right\} \right)\left[\prod_{j=1}^{n}\breve{\omega}_{\tau}^{k}\left(x_{j}\right)\right]dx_{1:n}\nonumber \\
 & \quad+\frac{p_{D}\left(x^{i}\right)\breve{\omega}^{k}\left(t,x^{1:i}\right)}{\ell^{k}\left(\mathbf{z}^{k}\right)}\sum_{z\in\mathbf{z}^{k}}l\left(z|x^{i}\right)\sum_{n=0}^{\infty}\left(n+1\right)\rho_{\omega^{k}}\left(n+1\right)\nonumber \\
 & \quad\times\int\ell^{k}\left(\mathbf{z}^{k}\setminus\left\{ z\right\} \left|\left\{ x_{1},...,x_{n}\right\} \right.\right)\left[\prod_{j=1}^{n}\breve{\omega}_{\tau}^{k}\left(x_{j}\right)\right]dx_{1:n}.\label{eq:PHD_CPHD_derivation1}
\end{align}
From \cite[Eq. (39)]{Angel15_d}, we know that 
\begin{align}
 & \sum_{n=0}^{\infty}\left(n+1\right)\rho_{\omega^{k}}\left(n+1\right)\nonumber \\
 & \quad\times\int\ell^{k}\left(\mathbf{z}^{k}|\left\{ x_{1},...,x_{n}\right\} \right)\left[\prod_{j=1}^{n}\breve{\omega}_{\tau}^{k}\left(x_{j}\right)\right]dx_{1:n}\nonumber \\
 & \quad=\int D_{\omega_{\tau}^{k}}(x)dx\left\langle \rho_{\omega^{k}},\Upsilon^{1}\left[D_{\omega_{\tau}^{k}},\mathbf{z}^{k}\right]\right\rangle \prod_{z\in\mathbf{z}^{k}}\breve{c}\left(z\right).\label{eq:preliminary4_cphd}
\end{align}
Substituting (\ref{eq:density_measurements_CPHD}) and (\ref{eq:preliminary4_cphd})
into the first and second term of (\ref{eq:PHD_CPHD_derivation1})
completes the proof of (\ref{eq:PHD_update_CPHD}).

\section{\label{sec:Appendix_TCPHD_prediction}}

In this appendix, we prove Theorem \ref{thm:TCPHD_prediction} (TCPHD
prediction). We proceed to calculate the cardinality and PHD of the
predicted multitrajectory density. As in the CPHD filter, the assumptions
of the TCPHD filter prediction (P1, P4, P5) are similar to the TCPHD
filter update (U1, U4, U5) so the proof of the TCPHD filter update,
which is given in Appendix \ref{sec:Appendix_TCPHD_update}, is useful
for the TCPHD filter prediction, as in the CPHD filter \cite{Angel15_d}. 

\subsection{Cardinality }

The cardinality distribution of the predicted multitrajectory density
for alive targets is analogous to the cardinality distribution $\rho_{\ell^{k}}\left(\cdot\right)$
of the measurement, which can be calculated from (\ref{eq:density_measurements_CPHD})
in the same way as in \cite[Sec. V.C]{Angel15_d}. The result is
\begin{align*}
\rho_{\ell^{k}}\left(m\right) & =\sum_{j=0}^{m}\rho_{c}\left(m-j\right)\sum_{n=j}^{\infty}\left(\begin{array}{c}
n\\
j
\end{array}\right)\rho_{\omega^{k}}\left(n\right)\\
 & \quad\times\frac{\left[\int\left(1-p_{D}(x)\right)D_{\omega_{\tau}^{k}}(x)dx\right]^{n-j}}{\left[\int D_{\omega_{\tau}^{k}}(x)dx\right]^{n}}\\
 & \quad\times\left[\int p_{D}\left(x\right)D_{\omega_{\tau}^{k}}(x)dx\right]^{j}.
\end{align*}
By changing $\rho_{c}\left(\cdot\right)$, $p_{D}\left(\cdot\right)$
and $\omega^{k}\left(\cdot\right)$ for $\rho_{\beta^{k}}\left(\cdot\right)$,
$p_{S}\left(\cdot\right)$ and $\pi{}^{k}\left(\cdot\right)$, respectively,
we obtain (\ref{eq:prediction_cardinality_CPHD}). 

\subsection{PHD}

The result for the PHD of the TCPHD prediction, which is the same
as for the TPHD prediction, see (\ref{eq:predicted_PHD}), can be
established directly based on thinning, displacement and superposition
of point processes, as in the CPHD filter for targets \cite{Angel15_d}.
Nevertheless, for completeness, we provide more details regarding
its calculation here. 

We have a multitrajectory density $\pi^{k-1}\left(\cdot\right)$ with
a PHD $D_{\pi_{k-1}}\left(\cdot\right)$ and we want to compute the
PHD of the multitrajectory density $\omega^{k}\left(\cdot\right)$,
which is $D_{\omega^{k}}\left(\cdot\right)$. We first compute the
PHD of $\xi^{k}\left(\cdot\right)$, which represents the multitrajectory
density of the surviving trajectories and is given by
\begin{align}
\xi^{k}\left(\mathbf{X}\right) & =\int f_{s}\left(\mathbf{X}|\mathbf{Y}\right)\pi^{k-1}\left(\mathbf{Y}\right)\delta\mathbf{Y}\label{eq:density_surviving_trajectories}
\end{align}
where $f_{s}\left(\cdot|\cdot\right)$ is the transition multitrajectory
density for surviving trajectories. We proceed to explain a decomposition
of $f_{s}\left(\cdot|\cdot\right)$, which will be useful for the
proof. 

We first write the probability of survival and single trajectory transition
density for alive trajectories at time $k$ as a function of trajectories.
These correspond to
\begin{align}
p'_{S}\left(t,x^{1:i}\right) & =p_{S}\left(x^{i}\right)\label{eq:prob_survival_trajectory}\\
g'\left(t,x^{1:i}|t',z^{1:i'}\right) & =\delta\left[t-t'\right]g\left(x^{i}|z^{i'}\right)\delta\left(x^{1:i-1}-z^{1:i'}\right)\label{eq:transition_density_trajectory}
\end{align}
where $t+i-1=k$ and $t'+i'-1=k-1$, otherwise, $p'_{S}\left(\cdot\right)$
and $g'\left(\cdot|\cdot\right)$ are zero. Also, $\delta\left[\cdot\right]$
and $\delta\left(\cdot\right)$ denote Kronecker and Dirac delta,
respectively.

We use the following decomposition for the transition density of surviving
trajectories
\begin{align*}
f_{s}\left(\left\{ X\right\} \cup\mathbf{X}|\mathbf{Y}\right) & =\sum_{Y\in\mathbf{Y}}g'\left(X|Y\right)p'_{S}\left(Y\right)f_{s}\left(\mathbf{X}|\mathbf{Y}\setminus\left\{ Y\right\} \right).
\end{align*}
Note that this decomposition results from the fact that given the
surviving trajectories $\left\{ X\right\} \cup\mathbf{X}$, trajectory
$X$ must have survived from a trajectory $Y$ that belongs to $\mathbf{Y}$
and the rest of the surviving trajectories $\mathbf{X}$ have survived
from the rest of the trajectories at the previous time step $\mathbf{Y}\setminus\left\{ Y\right\} $.

The PHD of the surviving trajectories is obtained by substituting
(\ref{eq:density_surviving_trajectories}) into (\ref{eq:PHD})
\begin{align}
D_{\xi^{k}}\left(X\right) & =\int\xi^{k}\left(\mathbf{X}\cup\left\{ X\right\} \right)\delta\mathbf{X}\nonumber \\
 & =\int\left[\int f_{s}\left(\left\{ X\right\} \cup\mathbf{X}|\mathbf{Y}\right)\delta\mathbf{X}\right]\pi^{k-1}\left(\mathbf{Y}\right)\delta\mathbf{Y}.\label{eq:appendix_surviving_PHD}
\end{align}
We simplify the inner integral in (\ref{eq:appendix_surviving_PHD})
as 
\begin{align*}
 & \int f_{s}\left(\left\{ X\right\} \cup\mathbf{X}|\mathbf{Y}\right)\delta\mathbf{X}\\
 & =\int\sum_{Y\in\mathbf{Y}}g'\left(X|Y\right)p'_{S}\left(Y\right)f_{s}\left(\mathbf{X}|\mathbf{Y}\setminus\left\{ Y\right\} \right)\delta\mathbf{X}\\
 & =\sum_{Y\in\mathbf{Y}}g'\left(X|Y\right)p'_{S}\left(Y\right).
\end{align*}
Substituting this equation into (\ref{eq:appendix_surviving_PHD})
and applying Campbell's theorem \cite[Sec. 4.2.12]{Mahler_book14}
\begin{align}
D_{\xi^{k}}\left(X\right) & =\int\left(\sum_{Y\in\mathbf{Y}}g'\left(X|Y\right)p'_{S}\left(Y\right)\right)\pi^{k-1}\left(\mathbf{Y}\right)\delta\mathbf{Y}\nonumber \\
 & =\int g'\left(X|Y\right)p'_{S}\left(Y\right)D_{\pi^{k-1}}\left(Y\right)dY.\label{eq:appendix_surviving_PHD2}
\end{align}
Substituting (\ref{eq:prob_survival_trajectory}) and (\ref{eq:transition_density_trajectory})
into (\ref{eq:appendix_surviving_PHD2}) yields $D_{\xi^{k}}\left(\cdot\right)$
in (\ref{eq:predicted_PHD}). By the superposition theorem of point
processes \cite{Stoyan_book95}, the PHD of $\omega^{k}\left(\cdot\right)$
is the sum of the PHDs of the surviving trajectories, which is given
by $D_{\xi^{k}}\left(\cdot\right)$, and the PHD of the new born trajectories,
which is given by $D_{\beta^{k}}\left(\cdot\right)$. Therefore, the
TCPHD prediction equation for the PHD is the same as for the TPHD
filter, see (\ref{eq:predicted_PHD}).
\end{document}